\documentclass[12pt,twoside]{report}

\newcommand{\reporttitle}{DeepHelp: Deep Learning for Shout Crisis Text Conversations}
\newcommand{\reportauthor}{Daniel Cahn}
\newcommand{\supervisor}{Dr. Ovidiu Serban \\Dr. Emma Lawrance}
\newcommand{\degreetype}{Computing: Machine Learning \& AI}

\usepackage[a4paper,hmargin=2.8cm,vmargin=2.0cm,includeheadfoot]{geometry}
\usepackage{textpos}

\usepackage{natbib} % for bibliography
\usepackage{tabularx,longtable,multirow,subfigure,caption}%hangcaption
\usepackage{fancyhdr} % page layout
\usepackage{url} % URLs
\usepackage[english]{babel}
\usepackage{amsmath}
\usepackage{amssymb}
\usepackage{graphicx}
\usepackage{dsfont}
\usepackage{epstopdf} % automatically replace .eps with .pdf in graphics
\usepackage{backref} % needed for citations
\usepackage{array}
\usepackage{latexsym}
\usepackage[pdftex,pagebackref,hypertexnames=false,colorlinks]{hyperref} % provide links in pdf

\hypersetup{pdftitle={},
  pdfsubject={}, 
  pdfauthor={},
  pdfkeywords={}, 
  pdfstartview=FitH,
  pdfpagemode={UseOutlines},% None, FullScreen, UseOutlines
  bookmarksnumbered=true, bookmarksopen=true, colorlinks,
    citecolor=black,%
    filecolor=black,%
    linkcolor=black,%
    urlcolor=black}

\usepackage[all]{hypcap}

\def\@makechapterhead#1{%
  \vspace*{10\p@}%
  {\parindent \z@ \raggedright \sffamily
    \interlinepenalty\@M
    \Huge\bfseries \thechapter \space\space #1\par\nobreak
    \vskip 30\p@
  }}

\def\@makeschapterhead#1{%
  \vspace*{10\p@}%
  {\parindent \z@ \raggedright
    \sffamily
    \interlinepenalty\@M
    \Huge \bfseries  #1\par\nobreak
    \vskip 30\p@
  }}

\allowdisplaybreaks

\graphicspath{{./figures/}}
\usepackage[symbol]{footmisc}

\usepackage[T1]{fontenc}

\DeclareMathOperator{\E}{\mathbb{E}}
\DeclareMathOperator{\B}{\mathcal{B}}
\DeclareMathOperator{\Yhat}{\hat{Y}}
\DeclareMathOperator{\yhat}{\hat{y}}
\DeclareMathOperator{\PP}{\mathrm{P}}
\DeclareMathOperator{\Var}{\mathrm{Var}}

\date{4 September 2020}

\begin{document}

% load title page
% Last modification: 2015-08-17 (Marc Deisenroth)
\begin{titlepage}

\newcommand{\HRule}{\rule{\linewidth}{0.5mm}} % Defines a new command for the horizontal lines, change thickness here

%----------------------------------------------------------------------------------------
%	LOGO SECTION
%----------------------------------------------------------------------------------------

\includegraphics[width = 4cm]{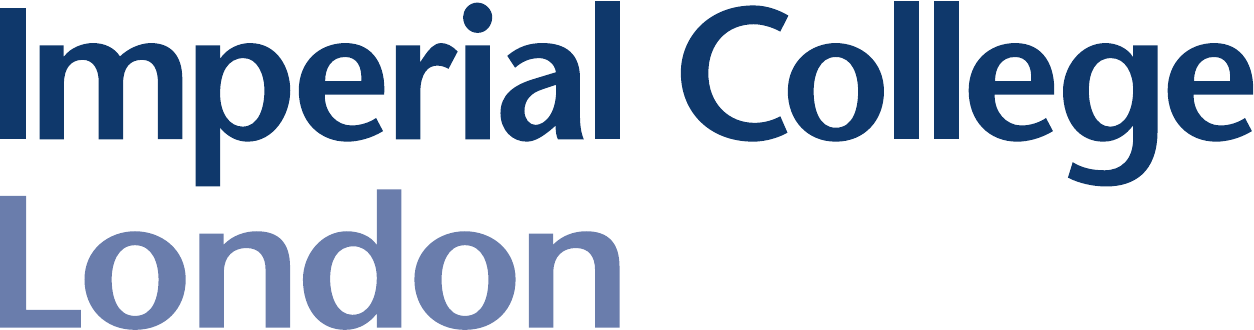}\\[0.5cm] 

\center % Center remainder of the page

%----------------------------------------------------------------------------------------
%	HEADING SECTIONS
%----------------------------------------------------------------------------------------

\textsc{\Large Imperial College London}\\[0.5cm] 
\textsc{\large Department of Computing}\\[0.5cm] 

%----------------------------------------------------------------------------------------
%	TITLE SECTION
%----------------------------------------------------------------------------------------

\HRule \\[0.4cm]
{ \huge \bfseries \reporttitle}\\ % Title of your document
\HRule \\[1.5cm]
 
%----------------------------------------------------------------------------------------
%	AUTHOR SECTION
%----------------------------------------------------------------------------------------

\begin{minipage}{0.4\textwidth}
\begin{flushleft} \large
\emph{Author:}\\
\reportauthor % Your name
\end{flushleft}
\end{minipage}
~
\begin{minipage}{0.4\textwidth}
\begin{flushright} \large
\emph{Supervisor:} \\
\supervisor % Supervisor's Name
\end{flushright}
\end{minipage}\\[4cm]

%----------------------------------------------------------------------------------------
%	FOOTER & DATE SECTION
%----------------------------------------------------------------------------------------
\vfill % Fill the rest of the page with whitespace

Submitted in partial fulfillment of the requirements for the MSc degree in
\degreetype~of Imperial College London\\[0.5cm]

\makeatletter
\@date 
\makeatother

\end{titlepage}

% page numbering etc.
\pagenumbering{roman}
\clearpage{\pagestyle{empty}\cleardoublepage}
\setcounter{page}{1}
\pagestyle{fancy}

%%%%%%%%%%%%%%%%%%%%%%%%%%%%%%%%%%%%

\begin{abstract}
The Shout Crisis Text Line provides individuals undergoing mental health crisis an opportunity to have an anonymous text message conversation with a trained Crisis Volunteer (CV). This project partners with Shout and its parent organisation, Mental Health Innovations, to explore the applications of Machine Learning in understanding Shout's conversations and improving its service. The overarching aim of this project is to develop a proof-of-concept model to demonstrate the potential of applying deep learning to crisis text messages.\\

Specifically, this project aims to use deep learning to (1) predict an individual's risk of suicide or self-harm, (2) assess conversation success and CV skill using robust metrics, and (3) extrapolate demographic information from a texter survey to conversations where the survey wasn't completed. To these ends, contributions to deep learning include a modified Transformer-over-BERT model; a framework for multi-task learning to improve generalisation in the presence of sparse labels; and a mathematical model for using imperfect machine learning models to estimate population parameters from a biased training set. \\

Key results include a deep learning model with likely better performance at predicting suicide risk than trained CVs and the ability to predict whether a texter is 21 or under with 88.4\% accuracy. Three metrics are produced for conversation success: how likely the texter is to have found the conversation helpful; how likely the CV is to generally have helpful conversations; and how much experience the CV seemed to have. The first achieves 90.0\% accuracy compared to texter survey responses while the second and third explain 25.2\% and 52.4\% of variance of the relevant ground truth labels. Finally, reversal of participation bias provides evidence that women, who make up 80.3\% of conversations with an associated texter survey, make up closer to 73.5\%- 74.8\% of all conversations; and that if, after every conversation, the texter had shared whether they found their conversation helpful, affirmative answers would fall from 85.1\% to 45.45\% - 46.51\%.

\end{abstract}

\cleardoublepage
%%%%%%%%%%%%%%%%%%%%%%%%%%%%%%%%%%%%
\section*{Acknowledgements}

Thank you to Dr. Mark Ungless, Dr. Ariele Noble, the Shout coaches and crisis volunteers I spoke with and the entire team at Shout and MHI for making the Shout dataset available for this research project, for providing a constant source of insights and suggestions, and for the incredible work that you do! \\

Thank you to Dr. Ovidiu Serban and Dr. Emma Lawrance for invaluable guidance throughout this project and for providing numerous ideas and suggestions which made this project possible. Thank you also to Dr. Lucia Specia for instruction in Deep Natural Language Processing and suggestions based on this project's progress report. \\

Thank you to Dr. Robert Peach and Dr. Tarik Altuncu for your collaboration on various aspects of this project.\\

Thank you to Edward Conway at MHI as well as Neil Clifford at the Imperial BDAU for helping me access the Shout dataset and the technologies needed to complete this project. \\

Thank you to my friends, and especially Michael, Mike, Gui and Alex, for helping me maintain my sanity throughout the Covid-19 lockdown. \\

Finally, thank you to my partner, Ashley, for everything.

\clearpage{\pagestyle{empty}\cleardoublepage}

%%%%%%%%%%%%%%%%%%%%%%%%%%%%%%%%%%%%
%--- table of contents
\fancyhead[RE,LO]{\sffamily {Table of Contents}}
\tableofcontents 

% \clearpage{\pagestyle{empty}\cleardoublepage}

\clearpage{\pagestyle{empty}\clearpage}

\pagenumbering{arabic}
\setcounter{page}{1}
\fancyhead[LE,RO]{\slshape \rightmark}
\fancyhead[LO,RE]{\slshape \leftmark}

%%%%%%%%%%%%%%%%%%%%%%%%%%%%%%%%%%%%
\chapter{Introduction}
\label{chapter:intro}

\section{Motivation}

The Shout Crisis Text Line provides individuals undergoing mental health crisis an opportunity to have an anonymous text message conversation with a trained Crisis Volunteer (CV). As Shout grows, its data is increasingly valuable as a new lens to understand how mental health issues affect individuals across the UK. By reading through anonymised conversations, an opportunity arises to learn the nature of mental health crisis by hearing directly from those affected, at scale, which can help elucidate the factors that can lead or contribute to crisis, and what can best help those in crisis. In parallel, as demand for the service grows more quickly than can be accommodated, a need is arising for computational solutions to ensure that conversations are at a consistently high quality. \\

This project will partner with Shout and its parent organisation, Mental Health Innovations, to explore the applications of Machine Learning for understanding Shout's conversations and improving its service. As the first research project on the Shout dataset, and likely the first major attempt to apply advanced deep learning to a crisis text messaging service, this project is a proof-of-concept demonstrating the potential in applying deep learning to crisis text messages. Specific aims of this project include using deep learning to (1) predict an individual's risk of suicide or self-harm, (2) assess conversation success by developing robust outcome metrics, and (3) extrapolate demographic information from a texter survey. \\

\subsection*{Suicide Risk}

During conversations that discuss suicide, CVs are required to perform a suicide risk assessment. This involves asking specific questions to understand the level of a texter's risk of suicide and reporting the results of the assessment on the Shout platform backend. This process is essential to ensure that high risk conversations are handled correctly and escalated, where necessary, to supervisors. Monitoring suicide risk can also save lives by triggering ``Active Rescues'' when absolutely necessary, in which authorities are called to intervene. A significant worry of some supervisors is risk that is missed, either through unsatisfactory risk assessment or a failure of a CV to report risk through the Shout platform. Machine Learning can be used to highlight conversations where high risk may be present and may have been missed by the CV. Additionally, predicting risk of suicide or self-harm is non-trivial and is thus an excellent test of a machine learning model's ability to understand conversations at a deep level, taking into account word usage, semantics, syntax, tone, and even emoji usage. \\

In addition to real-time suicide risk prediction, post-conversation analysis can be used to identify conversations where risk may have been missed, helping supervisors provide feedback to a CV that can improve their ability to assess risk in future conversations. Identifying these conversations can thus be a significant contribution to Shout's ability to audit CVs and ensure they are monitoring suicide risk sufficiently. Further, successful ability to predict risk retrospectively could allow for automatic labeling of a greater number of conversations, for use training a real-time model to detect risk. As will be discussed in chapter \ref{chapter:suicide_risk_ladder}, this is especially
important as current performance of suicide risk assessments may involve significant subjectivity which may introduce significant noise. 

\subsection*{Measuring Skill}
Machine Learning can also be used more generally to assist in the supervision of CVs. For quality control and for CVs to improve over time, supervisors and coaches consistently monitor and review a small subset of conversations for each CV and provide detailed feedback to help them improve. However, due to the great number of Shout conversations, it's impossible for supervisors to review every conversation and at present, the conversations that are reviewed are either self-selected by the CVs or selected randomly. Machine Learning can be used to identify conversations that went particularly well or poorly or that might contain specific issues for supervisors to review; this can improve supervisor's efficiency, allow for more useful feedback, and ultimately improve the Shout service. A robust conversation success metric can also be used as a label for a real-time model, which can be used to suggest supervisor intervention in conversations that take a negative turn. \\

Conversation success metrics are difficult to establish for every conversation. Prior to this project, the primary conversation success metric is a question in the texter survey asking whether a texter found their conversation helpful, and if so, how helpful. However, the texter survey is only filled out by a small number of texters with responses to the first question present in only 10.2\% of conversations analysed. As such, it is likely significant participation bias relating to conversation helpfulness, with texters who found a conversation very helpful or very unhelpful more likely to provide feedback. \citet{chamberlain2017give} find evidence for polarising participation bias - that those with very positive or negative experiences are more likely to review their experiences. They found that introducing an incentive to provide reviews significantly decreased the percentage of both one-star and five-star reviews, on a five-star scale. In the case of Shout, 37.6\% of conversation surveys with a helpfulness review gave a 5/5 helpfulness score (see figure \ref{fig:how_helpful}), a number which I later present evidence to indicate may not not be representative. \\

Beyond issues of participation bias, modelling texter feedback regarding conversation helpfulness might or might not directly reflect the success of a conversation from Shout's persepective. A texter's opinion on the helpfulness of a conversation might reflect more on the type of problem they're facing, their level of distress before entering the conversation, and whether the problem is one that can benefit from talking with a CV. It is therefore important to develop a robust success metric that can reflect the quality of the conversation or degree of skill shown by the CV, rather than merely the mental state of the texter. \\

A robust success metric may help Shout evaluate its own success, strengths and weaknesses; evaluate the strengths and weaknesses of individual CVs; understand how counselors improve over time as they gain experience; and perhaps identify when a counselors' success in conversations dips for a period of time. More generally, it might shed light on the factors that contribute to making a conversation successful or helpful. For example, what contributes most to making a conversation successful - counselor messages or texter responses? Does this vary by topic or demographics? Which stage of a conversation is most essential? \\ 

\subsection*{Survey Extrapolation}

Texter surveys are essential to understanding texter demographics, but might suffer from significant participation bias, i.e. where the same details that the survey captures also affect the likelihood that an individual opts in to the survey or answers specific questions in the survey. For example, \citet{hill2013wikipedia} found a major disparity between an opt-in study of Wikipedia users that estimated that 39.9\% of Wikipedia readers in the US were female and a slightly later study with more representative methodology by Pew Research Center, which found an even gender split. This could be explained by the finding in \citet{chang2009national}, which found that women are generally less likely to respond to opt-in Internet surveys than men. The effect of participation bias on surveys sent via text message, surveys relating to mental health, and specifically surveys from a crisis text line are less well-understood, and it is therefore essential to measure this effect in order to estimate true population demographics. In the case of Shout, a mere 15.8\% of survey respondents identify as male; 65.5\% as heterosexual/straight; and 64.9\% as 14-24 (see Chapter \ref{chapter:dataset}), bringing up a key question as to the degree of which these figures are representative.  \\

Deep learning can be used to estimate the degree of participation bias and counter it, to a limited degree, by extrapolating survey answers onto conversations without corresponding surveys. The extrapolation of demographic information will provide a better understanding of Shout's texter-base and shed light on various future research questions. Besides learning more about the full texter-base distribution over age, gender, and other variables, it can provide insight into the topics discussed by different populations; the times that those populations text in; and the presence of over and under-represented groups. Knowledge of these factors can also potentially better Shout's service, for example, by leading to further outreach to populations that text in less or by helping ensure that the population of CVs reflects well the population of texters. \\

\section{Objectives and Contribution}

In this project, I partner with Mental Health Innovations and their corresponding crisis text message service, Shout, to use Machine Learning to draw insights from two years of conversation data, including 10,808,136 messages from 271,405 conversations. The primary goal of this project is to develop a state-of-the-art deep learning model to ingest conversation text, provide meaningful outputs that can improve the Shout service, and provide output data that will be useful for the data scientists and clinicians that aim to better understand mental health crises. More specifically, this project will have three primary goals: (1) Extrapolate from texter surveys to the remainder of conversations to determine and partially reverse participation bias, especially for understanding trends related to demographic groups; (2) to develop a robust success metric for each conversation based on an estimate of the CV's skill; (3) to identify conversations where a CV misses, misidentifies, or insufficiently assesses risk of suicide or self-harm. \\

This project will have three primary contributions. The first is a contribution to deep learning including two novel methods of using the BERT model on sequences that exceed the length at which BERT can typically feasibly be used. These methods are Attention Over BERT, which involves combining multiple forward passes through BERT using an attention mechanism; and Transformer over BERT or ToBERT, which involves end-to-end training of hierarchical transformers, and is based on a model of the same name from \citet{hierarchicaltransformers}. Additionally, this project studies multi-task training as a form of semi-supervised machine learning in the presence of sparse labels. This includes a comparison of single-task training, multi-task training, and multi-task pre-training with single-task fine-tuning. \\

Second, this project will study the practical application of deep learning models to detect and reverse participation bias. This includes the development of a mathematical model for noise. This mathematical model is evaluated with numerical simulations, a small toy model trained on a biased subset of the MNIST dataset \citep{mnist}, and finally the practical usefulness of these methods is demonstrated through application to the Shout texter survey results. \\

And finally, this project contributes to the Shout service by identifying conversations to prioritise for review. This will be accomplished by identifying conversations that are found to be particularly successful or unsuccessful by three metrics, as well as where suicide risk was not sufficiently or correctly assessed. \\

Chapter \ref{chapter:dataset} of this report will explore the Shout dataset. Next, Chapter \ref{chapter:background} will provide a background on a number of NLP and Machine Learning techniques relevant to this project. Chapter \ref{chapter:implementation} will discuss the implementations of the models developed in this project, including increasingly complex models that allow for understanding the difficulty of a number of supervised learning tasks. Chapter \ref{chapter:evaluation} will evaluate each of the models and methods used in the previous chapter and discuss future model improvements. Chapter \ref{chapter:participation_bias} will develop a mathematical model for evaluating and reversing participation bias and will apply this model to the Shout dataset to generate insights on the true demographic spread. Chapter \ref{chapter:measuring_skill} will discuss methods for measuring CV skill using the machine learning methodology developed previously, analyse results, and discuss future research on this topic. Chapter \ref{chapter:suicide_risk_ladder} will discuss model predictions of suicide risk,  methods for analysis of insufficient suicide risk assessment, and will provide an analysis of results and ongoing work. Finally, chapter \ref{chapter:discussion} will provide a discussion of future avenues following the developments of this project, including the development of corresponding real-time models. \\

\section{Ethical and Professional Considerations}

Working with the Shout platform and dataset introduces a number of essential ethical and professional considerations. First, Shout texters must remain anonymous and conversations involve topics and information that are highly sensitive in nature. Ethically - and legally - it is important that this dataset be handled delicately to protect the privacy of the Shout texters. This project required training in GDPR and safe computing practices to ensure legal responsibilities could be respected.
Additionally, all data is only ever accessed through secure machines and secure connections. \\

It is also important that any personally identifying information in conversations such as names, phone numbers, and addresses is removed, or "scrubbed" from the dataset before it is used for research. This scrubbing process is performed by a team at MHI, but significant time was spent in this project assisting with improving scrubbing technology to ensure the data used was fully anonymised. It is also important that any information that slipped past the scrubbing process was immediately brought to the attention of relevant authorities to have it manually removed, as well as to better identify any systemic issues in scrubbing. Scrubbing reduces machine learning models' effectiveness due to over-scrubbing, especially where the scrubber confuses normal words with names or personally identifiable information. The removal of first names also significantly reduces the ability of models to predict an individual's race and gender. This is, however, a necessary sacrifice to protect texter privacy. Even after the data is fully anonymised, it is important that the dataset is carefully protected and that no part of it is ever made public as texters expect that their conversations are kept strictly between themselves and Shout.  \\

Additionally, the output of this project may significantly assist the Shout service. It is therefore essential to be aware that any machine learning models developed in this project are imperfect and should never be presented otherwise. When explaining my work to individuals at Shout, I've been careful to provide disclaimers on what the model can and cannot do to ensure it is never used as a substitute for human expertise. The models developed here are not intended to replace, but to supplement, humans.
\chapter{Dataset Exploration}
\label{chapter:dataset}
This project will involve the Shout dataset, a snapshot of 10,809,178 messages from 271,445 conversations spanning from 12 February 2018 through 3 April 2020, as well as various metadata related to messages, conversations, texters, and CVs. This chapter will provide an overview of the Shout dataset. Section \ref{section:dataset:conversations} will discuss the nature of Shout conversations as well as the preprocessing steps necessary to use these conversations for Machine Learning. Next, \ref{section:dataset:metadata} will discuss relevant metadata including the CV survey and the texter survey. Finally, \ref{section:dataset:analysis} will provide results of a preliminary data analysis using basic statistical methods, that was used to understand the dataset and inform the research to come. 

\section{Conversations}
\label{section:dataset:conversations}

Shout facilitates anonymous text message conversations between individuals in mental health crisis and Crisis Volunteers (CVs). Individuals become CVs by submitting an extensive application including professional references. The assessment of the applications aims to ensure, among other things, that CVs can be empathetic and non-judgemental. CVs then undergo a training process and progress through levels of seniority during which they gain increasing responsibilities and privileges in conversations, for example, allowing them to participate in two conversations at the same time. CVs are not paid and typically work in two hour shifts at least once per week. In addition to CVs, Clinical Supervisors oversee all conversations on the platform. They are clinically trained with appropriate psychological expertise. Coaches provide support to CVs off the platform, during training of CVs, and provide feedback and care for CVs' wellbeing after shifts. \\

Texters can learn about Shout through a number of avenues and can initiate a conversation at any time by texting to Shout's five-digit phone number. 
% Shout also has a number of partnerships with organisations including the NHS and The Mix in which the organisations advertise texting a special code (e.g. THEMIX) in initiating a conversation to indicate how they were referred. 
Upon initiating a conversation, texters receive an automatic machine reply enquiring what the texter is reaching out about, as well as informing them about Shout's privacy policy. Shout triages conversations based on these early messages to ensure that high risk messages are responded to especially quickly. In general, Shout aims to keep all wait times under five minutes, though wait times are generally longest in the middle of the night when the fewest CVs are active. \\

Once a conversation has been initiated, CVs are trained to follow five stages in a conversation: The 5 stages are: (1) Build Rapport (2) Explore (3) Identify the Goal (4) Discover Next Steps and (5) End the Conversation. Throughout a conversation, alerts may appear to the CV as pop-ups to remind them, for example, that they should have developed a rapport with the texter by this point, or that they should begin to wrap up the conversation. CVs are also required to assess risk of suicide on the Suicide Risk Ladder, if any part of the conversation suggests the texter may have some suicidal ideation; this will be discussed in more detail in Chapter \ref{chapter:suicide_risk_ladder}. If the risk is sufficiently imminent, an Active Rescue may take place, where authorities are called to physically intervene with a suicide attempt. CVs can also receive feedback or comments from a Clinic Supervisor during a conversation and conversations may occasionally be transferred to a Supervisor in circumstances where the CV requires extra assistance, for example in circumstances of high risk of suicide. \\

\subsection*{Preprocessing}

There are a number of normal pre-processing steps on conversation text data necessary for machine learning and data analysis. These include normalisation of characters, namely: converting emojis into their byte form; removing unicode characters such as those for underlining and italicising; converting both forms of dashes (``-'' and ``–'') into one form, and similarly with the three forms each of single and double quotation marks (e.g. `` and '' become \char34). Additionally, leading and trailing spaces are removed from messages and empty messages are removed. Finally, unicode characters that occur very infrequently, such as Chinese characters, are replaced with a special ``unknown'' token. \\

Regarding the dataset, it is worth mentioning that Shout publicly launched in May 2019, so the conversations between February 2018 and May 2019 are from Shout’s soft launch. Various early conversations are test conversations, sent in by Shout staff to test Shout’s system technically and for training purposes, though most of the conversations during Shout’s soft launch are genuine. \\

\subsection*{Conversation Length}
\begin{figure}[htb]
\centering
\includegraphics[width = 0.5\hsize]{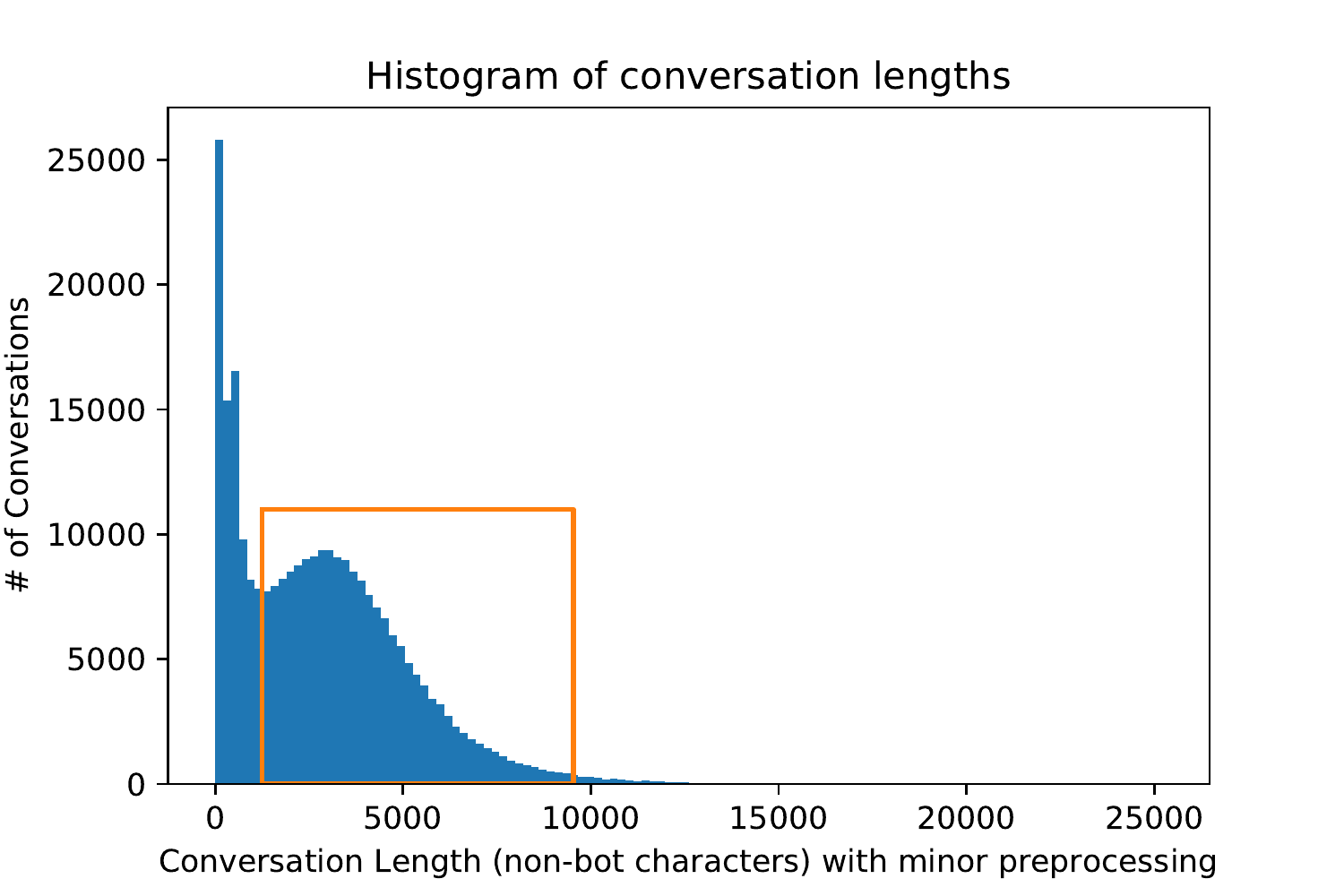}
\caption{Distribution of conversation lengths, and a box indicating the conversations used in analysis.}
\label{fig:conversation_lengths}
\end{figure}

Conversations vary significantly in length and duration. Measuring conversation length by total number of characters sent by both the CV and texter reveals a bi-modal distriubtion (see Figure \ref{fig:conversation_lengths}). After initial pre-processing, conversation length ranges from one character to 25,213 with a mean of 2852, median of 2582, inter-quartile range of 3351, and standard deviation of 2303. Individuals at Shout explain that texters often have very short conversations for a number of reasons. Sometimes, they wish to test the Shout system before reaching out for help, to understand exactly how Shout works – for example, that their communication is fully anonymous and private, that they’re communicating with real people, and that the service is free. Other times, individuals merely want to express themselves and are not looking for a conversation; they send one or a few messages just to get something off their chest and then end the conversation without needing a response. Conversations can also end early when wait times are high and it takes too long for the texter to be matched with a CV, especially during the night when the texter may have fallen asleep by the time they are matched. \\

Based on the bi-modal distribution and an understanding from conversations with Shout staff, it is clear that very short conversations might be outliers, containing less meaningful back and forth with a CV. Similarly, very long conversations will not be representative of the bulk of conversations. Due to the limited number of CVs, long conversations are discouraged, with warnings appearing to CVs when conversations have gone on too long, encouraging them to bring the conversation to a close. Nonetheless, in very heated situations or when there is imminent high risk and the texter should be encouraged to stay engaged, conversations can go on for hours. Instead of removing long conversations from the dataset, another option is to trim them, keeping only the first part of the conversation and simply not considering mes- sages after a certain point. An upside of this approach is that long conversations are more likely to be well labelled and might contain an important signal for a classifier. However, the stages of a conversation are likely to be quite different in long conversations, both because the conversations are likely to go through a greater number of stages and because stages will inevitably last longer. These conversations are therefore also unlikely to be representative of the bulk of the dataset.\\

Based on a visual analysis of the histogram of conversation lengths [figure], I remove conversations with length below 1,200. On the high side, I remove conversations with length greater than two times inter-quartile range above the mean. In total, 30.47\% of conversations were removed from below and 1.14\% from above. In total, 14.14\% of messages are removed at this step. \\

\section{Metadata}
\label{section:dataset:metadata}

\subsection*{CV Survey}

At the end of each conversation, CVs fill out a short survey with six parts:
\begin{enumerate}
\item Select which, from a list of issues, were discussed in the conversation. CVs have 15 options including ``Abuse, Physical'', ``Depression/Sadness'', ``Self-harm'' and ``Suicide''
\item If ``self-harm'' is ticked as an issue discussed: Select whether self-harm is in progress (yes, no, or not sure)
\item If ``suicide'' is ticked as an issue discussed: Suicide Risk Ladder assessment (checkboxes: Thoughts, Plans, Means, Timeline)
\item Write if any ``coping skills or safety plans`` that were agreed to by the texter (free response, multiple separate answers are allowed)
\item Write any ``signposts'' that were agreed to by the texter (free response, multiple separate answers are allowed)
\item ``How are you feeling?'' (``I’m good!'', ``Ehhh'', ``Upset'')
\end{enumerate}

\subsection*{Texter Survey}

Texters also receive surveys with several questions, with the instructions to fill out any that they feel comfortable answering, in any order. The total number of questions is roughly 25, though the questions that appear can vary depending on how earlier questions are answered. Questions have also varied slightly over time, for example, a Coronavirus question was added in early 2020. While the CV survey is filled out after nearly every conversation, the texter survey is filled out roughly 10.24\% of the time, based on the number of texters who answer the first question. Questions include some related to the conversation: \\

\begin{itemize}
    \item ``Did you find this conversation helpful?'' and if so ``How helpful was it?'' from 1 (slightly helpful) to 5 (very helpful)
    \item Open text area to leave a note for the counselor 
    \item Do they feel more, less, or equally ``in control,'' ``hopeful,'' ``alone,'' ``depressed,'' ``overwhelmed,'' ``upset,'' ``suicidal'' following the conversation
    \item Did they ``mention an experience or feelings that you have not shared with anyone else?'
    \item Do they ``believe my Crisis Volunteer was genuinely concerned for my well-being''? 
    \item Did they agree to a plan; if so, how likely they are to follow the plan; and if not, why not.
\end{itemize}

Other questions relate to the individual and demographics, including regarding their age (see Figure \ref{fig:age_dist}), gender (see Figure \ref{fig:gender_dist}), sexual orientation (see Figure \ref{fig:sexual_orientation_dist}), race, disabilities, and any past military experience. Texters are also asked about how often mental health issues were experienced in the prior two weeks, including feeling ``nervous or anxious'', ``worrying'', having ``little interest in doing things'', ``feeling depressed or hopeless'', ``feeling left out'', ``feeling isolated.'' In addition to texting in today, how else do they get help when in crisis? And finally, how did they hear about Shout?  \\

Texters can fill out the survey multiple times, but are not asked demographic questions on future occasions. Demographic responses are therefore generalised here across all conversations from the same phone number.

\section{Analysis}
\label{section:dataset:analysis}

\subsection*{Texter Survey Distributions}

\begin{figure}[htb]
\centering
\includegraphics[width = 0.5\hsize]{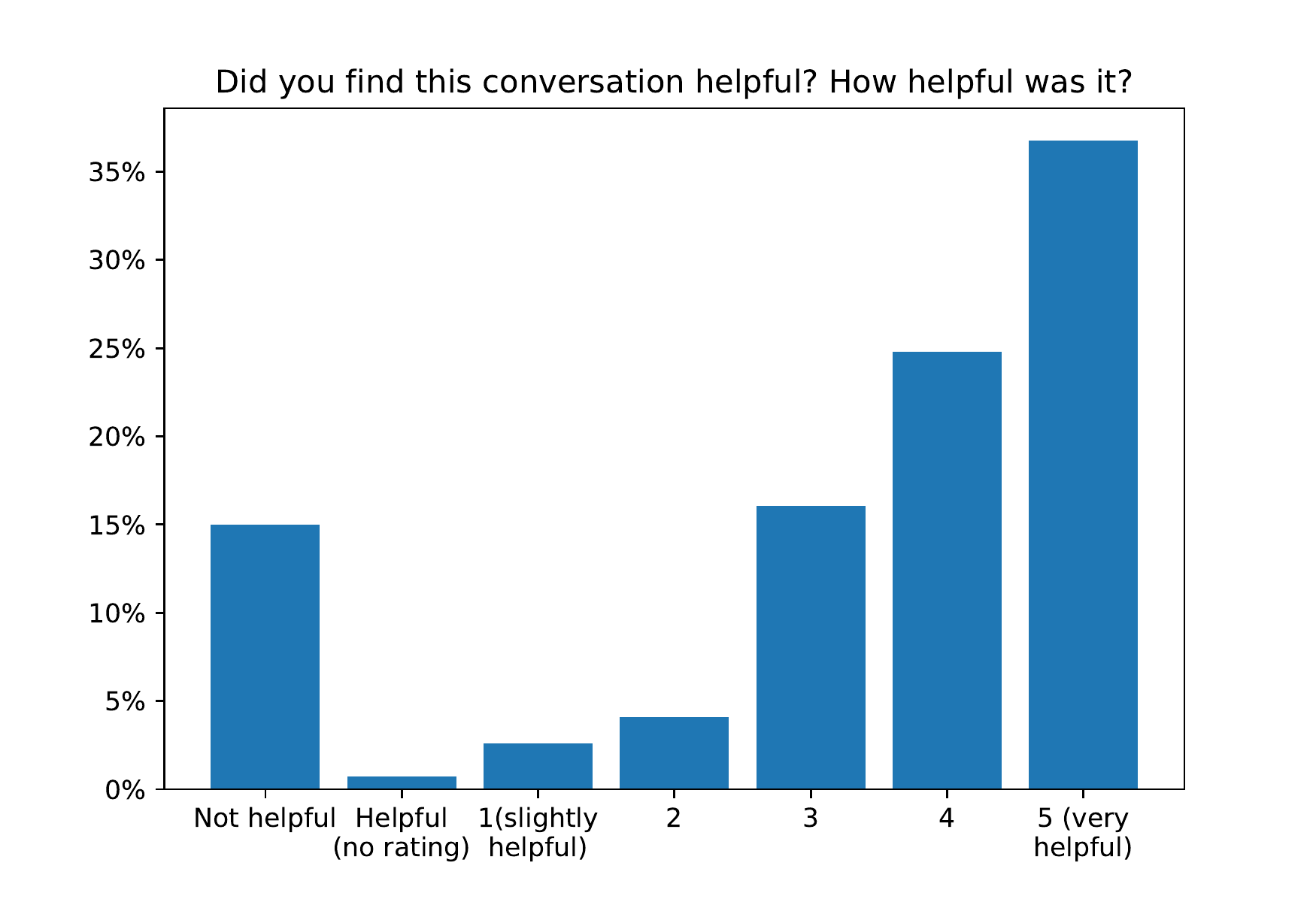}
\caption{Responses in the Shout texter survey for "Did you find this conversation helpful?" and "How helpful was it?".}
\label{fig:how_helpful}
\end{figure}

Due to the rarity of the texter survey being filled out, the survey results likely show significant participation bias; demographics from the texter survey can thus hardly be taken as representative of all texters. The conversation success metrics are also possibly skewed in a positive direction with 37.6\% of texters who responded to the question rating it 5/5 on helpfulness, (see Figure \ref{fig:how_helpful}). There is also significant sparseness to the survey answers, with texters only responding to a subset of the questions because all questions are optional.  There is also significant cross-correlation between survey questions, with some asking very similar questions with different wording. And there is likely to be noise, with different people interpreting questions differently, and of course different individuals considering a conversation ‘successful’ based on lots of subjectivity. \\

\begin{figure}[ht]
\centering
\includegraphics[width = 0.7\hsize]{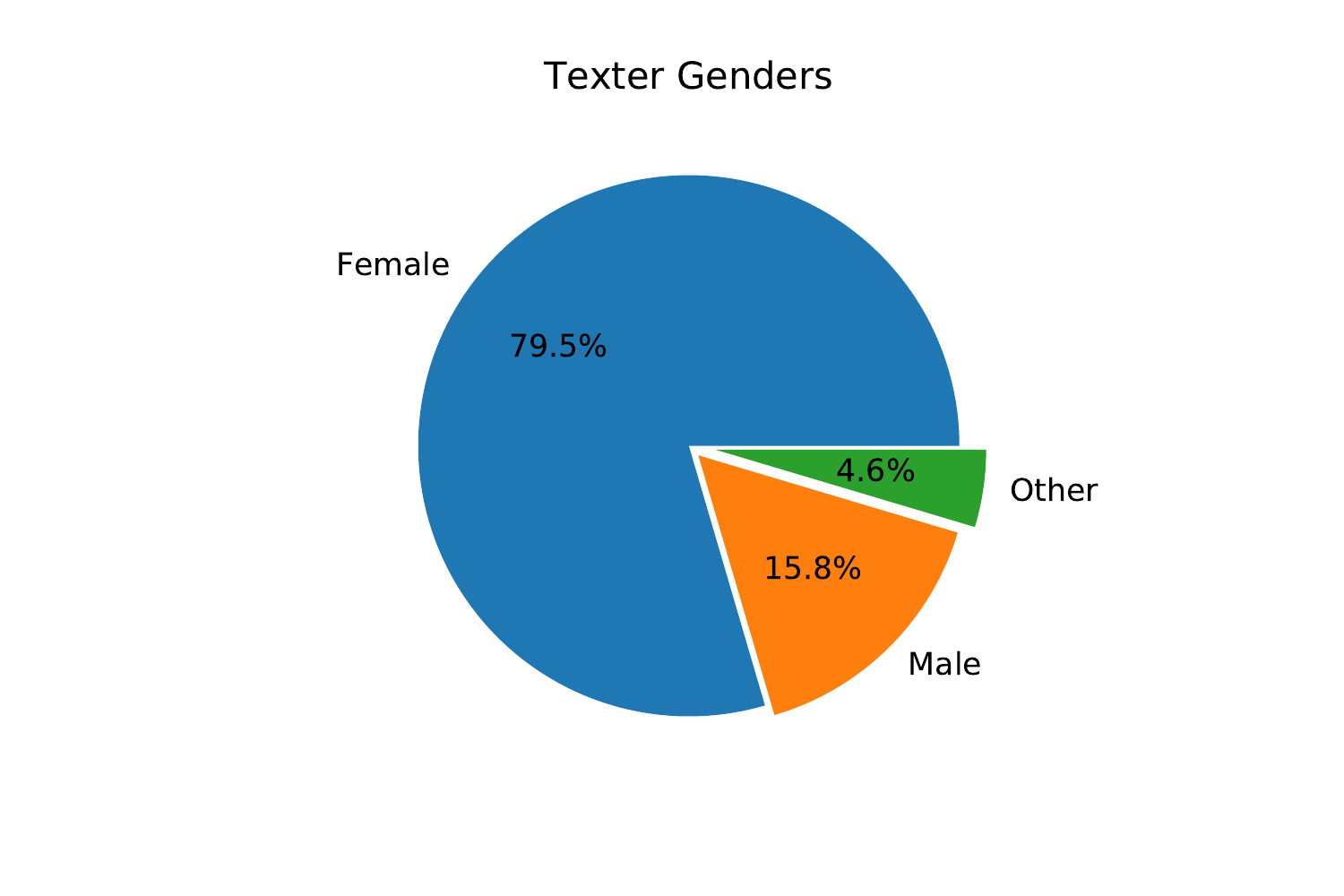}
\caption{Distribution of Shout texters' genders, as per the texter survey}
\label{fig:gender_dist}
\end{figure}

\begin{figure}[ht]
\centering
\includegraphics[width = 0.7\hsize]{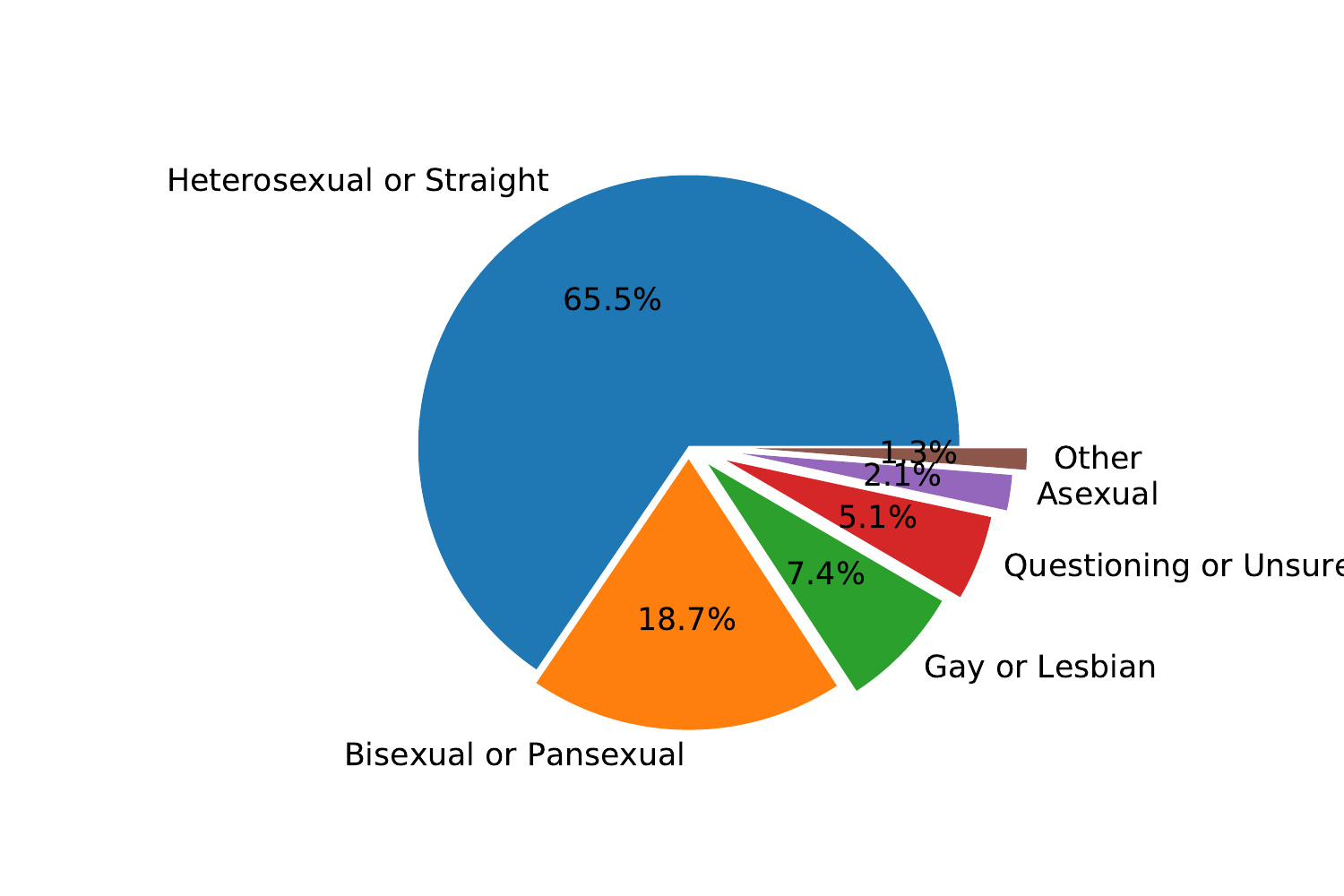}
\caption{Distribution of Shout texters' sexual orientations, as per the texter survey}
\label{fig:sexual_orientation_dist}
\end{figure}

\begin{figure}[ht]
\centering
\includegraphics[width = 0.5\hsize]{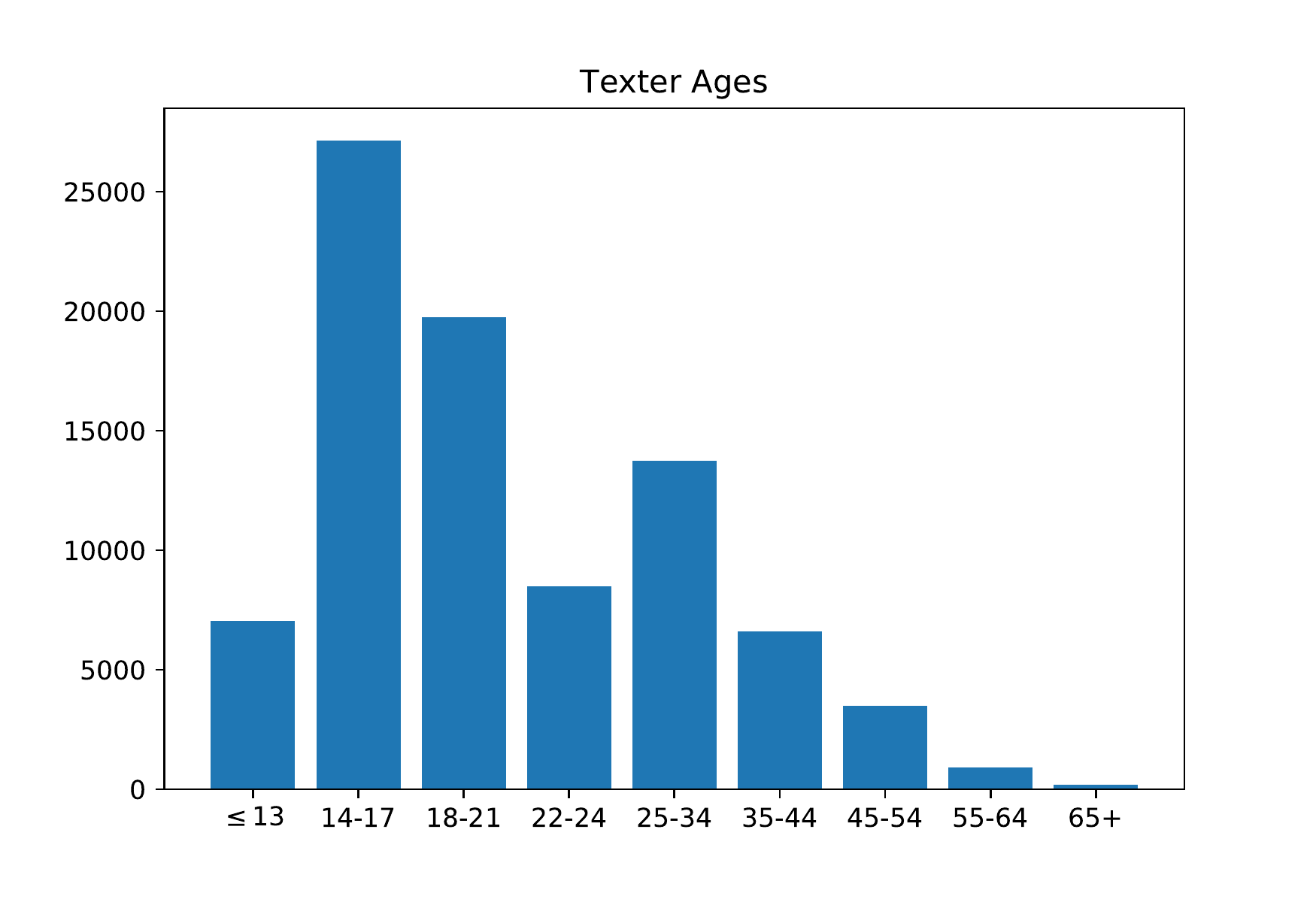}
\caption{Distribution of Shout texters' ages, as per the texter survey. Note that the bar widths do \textit{not} correspond with the size of the age ranges, do to the nature of the survey questions.}
\label{fig:age_dist}
\end{figure}

Additionally, based solely on survey data, it appears that females make up the vast majority of conversations (see Figure \ref{fig:gender_dist}), LGBT individuals make up nearly but not quite half of conversations (see Figure \ref{fig:sexual_orientation_dist}), and ages sharply peak among individuals age 14-21 (see Figure \ref{fig:age_dist}).\\

Looking at the correlations between word occurrences and specific labels, there are a number of words that can be used heuristically to differentiate one group from another. For example, helpfulness ratings are negatively correlated with texters using the word ``robot''\footnote[1]{$p\ll0.001$}, which may indicate that a significant complaint of texters may be CVs behaving too robotically. Those ratings are also positively correlated with texters using the word ``hope''*.  Identifying as under eighteen years old in the texter survey correlates with the use of abbreviations including ``ik''*, ``idk''*, ``ye''*, ``tho''* and ``rlly''*, a heuristic that may prove useful for CVs and for modelling age. Conversations related to substance abuse are distinguished by words like ``alcohol"*, ``cocaine"*, and ``sober''* - but are more interestingly separated by the presence of words like ``money"* and ``destructive"*. \\

Other simple factors may also be relevant in evaluating and distinguishing between conversations. Considering the helpfulness responses from the texter survey, one can see that conversations found helpful are longer on average than those found unhelpful. Texters in conversations that are found helpful write 81.56 more words on average* in 2.32 extra messages on average*, and CVs in these conversations write 58.10 additional words* and 2.90 additional messages*. CVs also on average write longer words in these conversation, by 0.054 letters per word*; texters write 2.08 more words per message*; and time between messages is 15.85 seconds less on average*.

\subsection*{Multi-Task Learning}

In this project, multi-task learning is used to improve model performance and ability to generalise due to  significant correlation between labels in spite of sparsity of relevant targets. Looking manually through conversation metadata, certain correlations are indicative of the need and value of multi-task learning; for example, the CV debrief question "How are you feeling?" at the end of conversations correlates with the texter survey question "How helpful did you find this conversation?" with a Matthews Correlation Coefficient (MCC) of 0.196*, on the conversations where both are present. While the CV debrief response is available in 96.9\% of conversations, the helpfulness response is available for only 10.2\% of conversations. A model trained simultaneously on both labels can capture the correlations between the auxiliary label (e.g. how the CV is feeling) and target label (how the texter is feeling), allowing for better modelling of the latter through modelling of the former.\\

In addition to modelling linear correlations between variables, training on auxiliary labels may cause a model to extract features from text that are likely relevant for training on target labels.  For example, an important free response question in the CV survey asks about any coping skills or safety plans agreed to by the texter. The presence of any answer to this question may correlate with how helpful texters find conversations, and different plans may interestingly correspond with different helpfulness reviews. In fact, plans that contain the word `talk' have the strongest correlation with texters finding the conversation helpful (MCC=0.067)* followed by those that mention `music' (0.067)* and `friend' (0.060)*, while those that contain `stop' are less likely to be found helpful (-0.038)*. It is clearly valuable for the model to capture correlations in these labels, but a more important effect of training on these labels is the effect on model features. A model trained to identify these coping skills and safety plans will likely model certain of the very same features that are useful in understanding how helpful a conversation is to a texter.\\

\chapter{Background}
\label{chapter:background}

I'll introduce document representation methods including the use of basic features, Bag of Words, TF-IDF, and word embeddings with Word2Vec and ELMo. I'll then present document classification models, including CNNs, RNNs, BERT, and document-level extensions to BERT models. Next, I discuss semi-supervised machine learning methods, to train on data with auxiliary labels. And finally, I discuss specific topics relevant to this project, namely estimating model uncertainty, which will be useful for predicting unsatisfactory risk assessments; and participation bias, which will be relevant to this project's work in identifying and reversing participation bias. \\

\section{Related Work}

Some research has been conducted previously into crisis-text lines, but this project might be the first to use deep learning and Natural Language understanding methods, and is the first to use the Shout dataset. \citet{althoff2016large} studies the differences between more and less successful counselors. They focus on factors like distances between TF-IDF encoded conversations that are going well and those that are going badly, conversation lengths, and lengths of messages. I, therefore, incorporate these in the baseline models to evaluate their discriminatory ability in predicting demographic and success-related measures. \citet{dinakar2015mixed} extracts conversation topics from messages using Latent Dirichlet Allocation, but does not predict conversation success or texter demographics and uses only a small dataset of 8,106 conversations. \\

Predicting user demographics based on text is the topic of some literature. \citet{predictingdemographicsmoocs} try to predict demographic information about users of an online course by applying NLP-methods on user comments. \citet{cesare2017machine} analyse  91 articles in the literature with methods of predicting age, race, ethnicity or gender from users of social media. The average accuracy across articles predicting gender was 83\%, however data used often included names, user profile information, profile pictures, and data matching with publicly available data - all methods that are either unavailable or unethical with an anonymised Shout dataset. \citet{iqbal2015predicting} try to predict twitter user's demographics based only on the authors' (non-anonymised) tweets, but achieve a mere 55.6\% accuracy in predicting gender on their perfectly balanced dataset, indicating the difficulty of the task. \\

\section{Document Representation Methods}

There are various methods for representing and classifying documents. To begin, there are simpler methods for representing a document by a feature vector that can be passed through a simple feature classifier such as a Logistic Regressor \citep{fan2008liblinear} or a Support Vector Machine SVMs \citep{joachims1998text}. The simplest feature set to represent a document is handpicked task-specific features from the document. Features can include, for example, the presence or frequency of specific task-dependent words or phrases used in a conversation by one or either speaker. However, it is often impractical to identify the best features for classification, and more advanced methods can be used for identifying features automatically.

\subsection*{Bag of Words / TF-IDF}

Features relating to the presence and frequency of word can also be obtained automatically with the bag-of-words model \citep{bagoftricks,zhang2015character, harish2010representationoftextdocuments}. The bag-of-words model represents a document as a vector with each index representing the number of times that a certain word appears in a document (or simply whether the word appears). Stemming is often performed as a preprocessing step, where the same word in different forms are mapped to a stemmed form (e.g. walking and walked are mapped to walk) for use in producing the resulting vector \citep{khan2010review}. Words that occur too frequently or not frequently enough may also be removed. \\

An addition to the Bag-of-Words model is the TF-IDF post-processing step, where the number of times a word appears in a document, or \textbf{T}erm \textbf{F}requency, is weighted by \textbf{I}nverse \textbf{D}ocument \textbf{F}requency, to represent the relative high importance of words that appear infrequently in a corpus but frequently in a single document, as well as the relative low importance of words that appear frequently across the entire corpus \citep{khan2010review}.\\

Bag-of-words and TF-IDF vectors have successfully been used to model topics and sentiment in a document \citep{joachims1998text, zhang2015character}, but have limited ability to model sequential nature of a text and inter-dependencies within a text. For example, the phrases ``I like it'' and ``I don't like it'' will be modeled very similarly, despite their opposite meanings. These methods also do not represent the many different meanings and connotations a single word can have. These methods can also be altered slightly to include bigrams or n-grams, i.e. n sequential words, to increase the ability to recognize phrases \citep{wang2012baselines}, however larger sequences will require a much larger corpus to capture interesting information) and the capacity for capturing relationship between slightly more distant words is still missing. 

\subsection*{Word2Vec}

The Word2Vec algorithm can be used to generate document representations with significant improvements upon bag-of-words and TF-IDF. Word2Vec refers to a method of generating vector representations of words on the assumption that the meaning of words can be represented by words with which they frequently co-occur\footnote{Many in the literature summarise this assumption with the words of \cite{firth1961papers} that ``You shall know a word by the company it keeps.''}. There are two prominent training processes to learn Word2Vec vectors, namely Continuous Bag-of-Words and Skip-gram; these are explained in detail in \citet{rong2014word2vec}. In sum, Word2Vec is trained by training a two layer neural network to predict a word based on its context (or a context word based on the target word) and extracting the intermediate layer as a word embedding. Word2Vec vectors, trained from a very large corpus, can then be used for other purposes such as document representation. Two prominent effects of Word2Vec vectors is that they encode synonyms and analogies. First, words with similar meanings tend to be encoded near each other; and second, relationships like ``King'' is to ``Queen'' as ``Man'' is to ``Woman'' are encoded in that, mathematically, the vector representations for each of these words have the relationship ``King'' - ``Man'' + ``Woman'' is very near ``Queen'' \citep{rong2014word2vec}. \\

Word2Vec can be used to encode a document by first converting each word to their corresponding pre-trained vector representation, and then combining the word vectors, for example by taking the mean. Because Word2Vec is trained on a very large corpus, it results in far fewer words being removed in the vectorization process for being unknown compared e.g. to TF-IDF. It maps words with closely related meanings near each other, which reduces redundancy that occurs in bag-of-words vectors, and does not require stemming, thus maintaining more subtle meaning in words. The resulting mean vector might therefore be more meaningful than a TF-IDF vector. However, Word2Vec still has various shortcomings in representing a document. Like TF-IDF, Word2Vec cannot encode which meaning of a word is being used, when a word has multiple meanings, or the connotation of the word in its context. Taking the mean of Word2Vec vectors also loses all meaning in a document encoded sequentially, as in the example above. 

\subsection*{ELMo}

Embeddings from Language Models, or ELMo, is an improvement on Word2Vec that can encode a word's meaning in its context, taking into account connotation in a way that Word2Vec cannot \citep{elmo}. ELMo encodes context-sensitive word features by using a Recurrent Neural Network language model that's pre-trained on a large corpus to predict the next word given all preceding words in a sentence, as well as the previous word in a sentence based on right context. The internal representations in each direction are concatenated for each word to form a contextualized word vector, which can be useful especially when context affects a word's meaning, or is necessary to disambiguate a word with multiple meanings. ELMo's ability to encode context features led to state-of-the-art results on various benchmarks, when it was released \cite{elmo}. \\

\section{Document Classification Models}

While vector-based methods try to encode a document into a fixed-size vector, typically regardless of the task at hand, a number of models exist to generate predictions based on an entire document, training document representation at the same time as training the classifier. I'll discuss four such model-based methods for document representation, namely attention, Convolutional Neural Networks, Recurrent Neural Networks, and Transformer/BERT models. It is important to note that these methods can be, and often are, combined in different ways.

\subsection*{Attention}

Word embedding models such as ELMo and Word2Vec do not easily provide sequence-level embedding. The attention methods allows for combining embeddings to include the most important features from each embedding. Attention involves passing an embedding vector through a neural network layer that outputs a score for the relative importance of that vector. The score for all words in the document are then put through a softmax function to map them to the range [0,1] with a sum of 1. A weighted mean is then taken of the word vectors, with the softmaxed scores as weights. This weighted mean can then be used as a document representation for classification, and the attention layer can be trained together with the final layer. A slight improvement on this algorithm takes the weighted average over \textit{features}, rather than over vectors, by calculating an attention score separately for each feature (i.e. for 300 feature word vectors, the attention layer has in-dimension and out-dimension both as 300). This method allows for incorporating the relatively most important features from each embedding into the document vector representation used for classification. This is different than vector-based methods because the attention layer is trained jointly with the classification layer or layers.  \\

\subsection*{Convolutional Neural Networks}

A Convolutional Neural Networks or CNN is composed of convolutional layers, that apply weights and biases to inputs in a moving window fashion. A document can be represented as a sequence of word vectors, where each word is represented by a learned embedding or pre-trained (e.g. by Word2Vec) word embedding. CNNs can also be applied at the character level, as in \citet{zhang2015character}. The convolutional layer applies a number of kernels (weights and biases) to a fixed of adjacent embeddings at a time, and outputs a single number; for example, for a 100 word text, a window-size of 3, and an embedding size of 300, the layer's kernel shape will be 3x300 and will map the first three words to a single number. The full output sequence will be 98 numbers, with the left- and right- most words not having the kernel applied with them in the center. A layer is composed on a number of these kernels, with each kernel outputting a specific feature from the text. In addition to the usual activation functions (e.g. ReLU) in between layers, pooling layers, especially max-pool layers, can also be used to extract the most important features from a sentence region to approximate position invariance and to increase the receptive field of the network and require fewer layers to reduce the whole document to a small representation. See, e.g. \cite{liu2017deep}.\\

CNNs are especially useful for identifying the presence of specific terms or phrases that are helpful for the classification task, though they have a few limitations. CNNs can require many layers to convolve text to a small enough representation for classification. With varying lengths of documents, the length of a CNN output can also vary, and thus must still be combined in order to get a fixed-length document representation vector for final classification. CNN outputs can be also averaged, max-pooled, or combined with the attention mechanism, but in any case with a loss of information. \\

\subsection*{Recurrent Neural Networks}

Recurrent Neural Networks or RNNs can process varying length sequences by reusing the same RNN cell with the same weights for each token. The RNN cell stores information from a cell in memory, and then uses that memory in processing the next token; they can therefore model sequential information in a way that the above mentioned models cannot. RNNs can map a sequence to a fixed-sized vector by simply outputting the hidden state (memory cell) after the final token in the sequence, or by combining all hidden states by taking the mean, element-wise or feature-wise max, or by using the attention mechanism. Gated RNNs such as Long Short Term Memory (LSTM) and the Gated Recurrent Unit (GRU) are improvements on the RNN that solve the vanishing gradient problem \citep{lstm,GRU_empirical2014}. While LSTMs are ``strictly stronger'' than GRUs \citep{GRU_empirical2014}, but GRUs have fewer weights, can therefore be trained more quickly, and thus generally perform comparably to LSTMs on language tasks \citep{lstm2018}. Additionally, Bi-directional RNNs, like ELMo, allow for modelling causal information in the backwards direction as well, which makes it easier to encode information from the start of a sequence that might be lost by the end\citep{bidirectionalrnns}. RNNs can also be multi-layer allowing for modelling more complex causal relationships. RNNs have also been shown to be especially successful in combining the outputs of a CNN \citep{cnnrnn, C_LSTM}.

\subsection*{Transformers}

The state of the art approach to representing and classifying texts is the use of Transformer models, and especially the \textbf{B}idirectional \textbf{E}ncoder \textbf{R}epresentations from \textbf{T}ransformers, or BERT, model. The original Transformer model \citep{attentionisallyouneed} was composed of a set of stacked identical encoders which map a sequence to a sequence embedding followed by decoders which produce the desired output - typically a translation. BERT uses only the encoder to encode text sequences for token-wise or sequence-wise classification.\\

The main feature of Transformers are self-attention layers, which are similar to the attention mechanism used above, and map a sequence of embeddings to a sequence of the same shape based on the relationship between each of the words in the sequence. Each word is projected, via a linear neural network layer, into key, query, and value vectors. The dot similarity between one word's key and another word's value serves as an attention score for the second word with respect to the first. The softmax of these attention scores is taken and used as weights to form a weighted mean of all embeddings, to serve as a representation of a single word. By using a number of ``attention heads'', each composed of linear layers to project keys, queries, and values, features can be extracted about words based on their sequence. In addition to self-attention layers, transformers use linear layers, non-linearities, LayerNorm layers, skip connections, and other standard elements of modern neural networks. Importantly, self-attention layers do not directly make use of the sequential nature of text, treating all input embeddings equally in the attention step. It is therefore necessary to supplement word embeddings with positional embeddings, vectors added to each word embedding depending on its position in the sequence, to ensure that position can be taken into account. The main benefit of Transformers over RNNs is their ability to model sequence dependencies that are far more distant than RNNs \citep{attentionisallyouneed}. \\

The BERT model uses transformers as well as a protocol for two-part training to achieve state of the art results in several NLP benchmarks related to token as well as sequence classification\citep{bert}. To encode words in their context, BERT training involves masked language modelling, where a random subset of words in a sequence are substituted with a [MASK] token, and the model is trained to predict the masked word\footnote{Some words are replaced with random words or kept the same, to develop embeddings for all words, not merely masked words}. To learn sentence level representations, BERT simultaneously trains an embedding of whether two adjacent sentences actually follow each other. In training, two sentences that half of the time follow each other are, half of the time, are random two sentences, are passed together through BERT separated by a special [SEP] token and preceded by a special [CLS] token. In addition to position embeddings, segment embeddings are added to tokens from each sequence to help the model distinguish between segments from each sentence in the masked language task. The [CLS] token is then trained to encode whether the two sentences are actually sequential\footnote{There is no need to combine the embeddings for this sequence-level prediction because the attention mechanism already present within the Transformer model.}. This training process is called pre-training, and is done on a very large corpus to train BERT to model a language (or multiple languages). BERT is then fine-tuned for a specific task by maintaining the trained weights but removing the final layer and replacing it with a task-specific layer. For sequence representation tasks, the [CLS] token embedding is used as a fine-tunable feature representation of the sequence, and for token classification, each token's corresponding output is treated as a token embedding. These embeddings can be passed through a single linear layer to perform classification. \\

Since the release of the original BERT model, various improvements have been produced of which this project will make use of two. RoBERTa, or a Robustly Optimized BERT Pretraining Approach reuses BERT's exact model architecture but changes hyperparameters in the pre-training process and results in significant improvements in NLP benchmarks \citep{roberta}. Two important changes to BERT in RoBERTa are the removal of the next sentence prediction task and the use of 512 tokens from a document instead of just two sentences. RoBERTa was found to be even better at sequence-level NLP tasks despite the removal of this pre-training task. Importantly, RoBERTa can therefore be quite useful for modelling tasks such as document classification, where the number of tokens in a document is frequently greater than 512. Two other results of these changes are that the [CLS] token is not used in pre-training, and therefore is not especially useful for classification in fine-tuning; and additionally, segment embeddings are not pre-trained because several segments may be present in the text. \\

In addition to improvements to the pre-training process, a recent addition to the BERT space are the advent of distilled versions of BERT models. Distillation \citep{DoDeepNetsReallyNeedToBeDeep, hinton2015distilling} refers to the process of transferring knowledge learned by a ``teacher'' neural network, to a small ``student'' network. In \citet{distilbert} a pre-trained BERT model has its knowledge distilled into a much smaller network called DistilBERT. DistilBERT was found to still be fine-tunable and retain 97\% of BERT's performance despite having 40\% fewer weights and being 60\% faster. More recently, the same technique has been brought to RoBERTA, in a model so-called DistilRoBERTa. \\

\subsection*{BERT for Documents}

Because BERT takes all tokens into account at once, and the computational time and memory is $O(N^2)$ in sequence length, there is a limit to the size of sequences that can realistically be encoded at once through BERT. Several methods have recently been proposed for handling documents longer than BERT model's maximum token length, which is typically 512 token, however research is still ongoing\footnote{``To the best of our knowledge, no attempt has been done before to use the Transformer architecture for classification of such long sequences.'' \citep{hierarchicaltransformers}}. \citet{docbert}, for example, use knowledge transfer to train an LSTM to achieve similar performance to an RNN in a model called DocBERT. They only, however, experiment with sequences truncated to 512 tokens. \citet{transformerxl} reintroduce the notion of recurrence with Transformer-XL, a model that passes a Transformer sequentially on small segments of a document, sharing states between each segment pass, but not propagating gradients between segments. Very recently \citet{longformer} replace BERTs self-attention layers in their Longformer model, from the original version with quadratic complexity in sequence length to one that scales linearly. However, Transformer-XL and Longformer are both auto-regressive, which is not optimal for sequence classification where bi-directional features representations may be important. The extremely recent Big Bird model \citep{bigbird} maintains BERT's core structure while removing BERT's quadratic time and memory requirements in sequence length, but is still in its early stages.\\

\citet{hierarchicaltransformers} develop two methods for simplying applying the normal BERT model on longer sequences, namely Recurrence over BERT, RoBERT, and Transformer over BERT, ToBERT. The former extracts features from a sequence using a pre-trained-only BERT and then passes the feature representation of the text through a small RNN. In this paper, the authors extract features from BERT by pass 200 tokens at a time through BERT, with an overlap of 50 tokens between runs. ToBERT, works similarly but uses a Transformer instead of an RNN to process the BERT-encoded sequence. RoBERT and ToBERT achieve similar results in experiments, with ToBERT performing slightly superior on some. Neither involve fine-tuning of BERT and exact details of implementation are not provided in the paper; adjustments to this model are therefore made in the implementations for this project. Hierarchical BERT, HiBERT, is similar to ToBERT, except that it encodes sentences with a first transformer rather than segments, and is designed for document summarisation \citep{hibert}.  \\

%%%%%%%%%%%%%%%%%%%%%%%%%%%%%%%%%%%%

\section{Model confidence and uncertainty}
In the context of risk assessment, it is valuable to understand not simply an individual's predicted risk but the model's confidence or uncertainty in its risk assessment. In this project, model confidence is relevant for at least two tasks. First, highlighting risk missed by a CV in accordance with model confidence that that this risk is present; and second, identifying when risk assessment contains high uncertainty, indicating a CV's failure to assess risk sufficiently.\\

There are several neural networks methods to estimating uncertainty. Bayesian Neural Networks, or BNNs, for example, use distributions instead of normal weights to predict directly a posterior distribution \cite{mackay1992practical, hernandez2015probabilistic}. BNNs, however, require a greater number of weights and computational cost, and are therefore not feasible for this project. Importantly, \citet{gal2015dropout} find that the use of Dropout, where weights or neurons are randomly set to zero, serves as an estimation of Bayesian uncertainty. They find that Dropout, which is typically used in training to develop more robust models that overfit less \citep{srivastava2014dropout}, can be left on at test time to approximate uncertainty by performing a number of forward passes with different neurons or weights turned off, and treating the distribution of outputs as a sample of the distribution over possible models. They also find, empirically, that Dropout provides a reasonable uncertainty estimate for classification with the LeNet \cite{lenet} model on the MNIST \cite{mnist} dataset. \\

Models can also output confidence estimates directly. Consider a classifier that predicts whether a certain example is a member of a certain class by outputting a number between 0 and 1 that corresponds with the confidence of the example belonging to the class. Accuracy is defined with a decision boundary, typically 0.5, assigning a classification to each example based on whether the model's output is above or below the decision boundary, and then simply counting the fraction of examples that are correctly classified. A highly accurate model may be poorly calibrated, though, if the model's outputs don't correspond well with the probability of the example belonging to the class. A well calibrated model should be one where, for example, where 80\% of examples with a classification output near 0.8 are classified correctly. One way to measure the degree that a model is well calibrated is the Expected Calibration Error \cite{naeini2015obtaining}, or ECE, which is the expectation of the difference between the model's confidence and accuracy. To calculate this, one first places examples in equally spaced bins based on the model's confidence for that example; then one takes the absolute difference of the model's accuracy in that bin and the model's average confidence for that bin; and finally, the weighted average is taken of each bin's calibration error, weighted by number of examples in each bin. \\

Treating ECE as an estimate for calibration error, \cite{guo2017calibration} find that deep neural networks are typically more highly accurate at the cost of being badly calibrated. They evaluate a number of post-processing calibration methods, including histogram binning, isotonic regression, Bayesian Binning, and Platt scaling and find variations on Platt scaling to generally perform significantly better than the rest. To perform Platt scaling on a trained neural network, one simply freezes the weights of the network, takes the model logit outputs (prior to the Sigmoid activation), and uses them to train a logistic regressor on a validation set. 

\section{Multi-task Learning}
In this project, a key to performing successful supervised learning in the absence of a large quantity of labeled training data will be to make use of unlabeled training data and auxiliary labels with multi-task or multi-label classification.\\

For models that are not deep neural networks, performing multi-label classification is non-trivial, as most models can only classify in a binary manner, and using a set of models to classify each label separately - a method called Binary Relevance - does not benefit from cross-correlations of labels \citep{zhang2013review}. Various alternatives have been proposed, such as training only on the labels that are most representative \citep{bi2013efficient}; using label power-sets, where each possible subset of labels is treated as one label; the use of Classifier Chains; and various others \citep{dembszynski2010label, balasubramanian2012landmark, hsu2009multi, zhou2012compressed, yu2014large, bhatia2015sparse}. However, with Deep Neural Networks, one can simply have a final layer output label predictions that are trained simultaneously, and the internal representation of features will be taken into account in the model's intermediate layers \citep{zhang2006multilabel}.

Even with deep neural networks, however, when the label-space is large, it has been suggested that the labels can be treated as a high dimensional vector in label-space, rather than independent labels \citep{chen2012feature}. Label space can then be compressed to a smaller, more dense, dimensionality that can be more easily trained upon, and where label cross-correlations will be better taken into account \citep{tai2012multilabel}. A straightforward approach to this is to perform one of many variations on Principle Component Analysis to condense the label space to reduce the dimensionality, or Label space dimension reduction \citep{chen2012feature}. However, such methods do not handle well missing labels, which often occur in the Shout dataset. \citet{yeh2017learning} suggest a deep auto-encoder architecture to encode non-linear relationships between labels, and to allow for training with missing labels; they find positive results with their Canonical Correlated Autoencoder (C2AE) with up to 50\% of labels missing.\footnote{See also \citet{deeplearningmultilabel}} More recently, however \citet{liu2017deep} find Binary Cross Entropy applied to all labels to work sufficiently well on datasets with between 103 and 501,069 labels. \\

\citet{multilabelneuralnetwork} find that a simple Neural Network trained on multiple targets with cross entropy loss can exploit correlations between the various target. \citet{duan2016learning} find success in the presence of noisy labels by training on auxiliary less-noisy labels. \citet{collobert2008unified} find significant benefit in semi-supervised multi-task learning for NLP tasks. They train a word-level CNN to simultaneously predict ``part-of-speech tags, chunks, named entity tags, semantic roles, semantically similar words and the likelihood that the sentence makes sense.'' Their training methodology simply involves choosing a task and batch randomly in each training iteration, so that different task-specific datasets can be used for each task. \citet{mtdnn1} similarly find significantly improved results with multi-task training on NLP tasks. In both of these, it is understood that deep neural networks are able to learn better feature representations of text through multi-task training, and that selecting a random task for loss calculation roughly approximately minimising a loss composed of the sum of task specific losses. \citet{mtdnn2} improves \citet{mtdnn1} by adding BERT as the shared texter encoder, and successfully provides benefit from multi-task training with BERT. They also suggest that multi-task training ``profits from a regularisation effect via alleviating overfitting to a specific task, thus making the learned representations universal across tasks.'' \footnote{See also \cite{guo2018soft}}\\

In \cite{t5}, a model appropriate for multi-task training is developed called Text-to-Text Transfer Transformer, or T5. In training T5, the authors converts various GLUE \citep{glue} NLP tasks, such as summarisation, sentence acceptability judgment, sentiment analysis, question answering, and others into text-to-text problems, to then on all of these various tasks. Similar to BERT, they first pre-train the model on an unsupervised language modelling task; however by converting all of the GLUE tasks to a single form, the model can then be trained simultaneously on multiple tasks. The authors generally find benefit from pre-training with a multi-task objective and fine-tuning on a specific task, but find that different variations of training procedure produce superior results on different benchmarks. Both \cite{mtdnn2} and \cite{t5} find benefit from fine-tuning to specific tasks after multi-task pre-training. They also discuss problems they encounter with training a single network for multiple tasks, related to overfitting or underfitting to particular tasks due to having different ratios of data available for training each task, the difficulty of regularising the network appropriately for each task, and the fear of task interference ``where achieving good performance on one task can hinder performance on another'' \citep{t5}. \\

\section{Participation Bias}
Much of Shout's understanding of texters comes from the texter survey, which may suffer from significant participation bias. Participation bias, also sometimes referred to as non-response bias or self-selection bias refers to the bias in survey answers due to the variation in different groups' propensities to respond to the survey \citep{fowler2013survey}. Participation bias is a well studied topic with many in the literature interested in understanding its degree, the specific factors that affect propensity to respond to surveys, and efforts to reverse its effect. As discussed earlier, \citet{hill2013wikipedia} find gender bias in an opt-in internet survey and \citet{chamberlain2017give} find that those with very positive or negative reviews to share are more likely to leave reviews. \citet{telephonesurvey} find that individuals who initially refused to participate in telephone survey were ``older, were less likely to be currently married, were less likely to have a managerial occupation, had fewer lifetime sexual partners, and were more likely to have a history of diabetes.'' \\

These studies are just a few of many handling participation bias, but the research into response bias can hardly be used to assume any specific bias in the Shout texter survey. Because Shout's texter survey is with individuals in mental health crisis, because the texter survey is sent over text message, and for a number of other reasons, Shout's texter survey response bias is likely unique and cannot be understood by simply generalising from research into other surveys. There are, however limited methods for addressing participation bias in the literature. \\

One method that has been proposed for reducing the effect of participation bias is Propensity Score Adjustment (PSA) \citep{rosenbaum1983central}. PSA involves estimating an individual's likely to respond to a survey based on covariates such as demographics, and then reweighing survey responses based on these propensity scores. In \citet{ferri2020propensity}, simple machine learning algorithms are used to determine the propensity scores used to re-weight survey responses in web surveys. However, PSA relies on information about coviarates in the underlying distribution; for example if gender were used as a covariate, it would be necessary to estimate the gender distribution across all Shout texters. Most of the covariate information is unavailable in the case of Shout and in many comparable use cases due to the strict anonymity of the service where randomised follow-up is simply not possible. \\

The most useful co-variate in the Shout dataset is not an auxiliary label but the text of the conversations themselves. With a limited degree of accuracy, conversation text can be used to predict answer to many texter survey question such as age, gender, and how helpful texters find a conversation, as will be shown in the sections to come, deep learning therefore makes possible more useful methodologies than re-weighing survey results based on propensities; namely, models can be used to predict the texter survey answers for all conversations and model predictions can be used to extrapolate from texters surveys and predict population estimates directly from conversations. \\

Further, the use of masked language modelling enables the model to train and extract features based on the full distribution of conversations; this is expected to help the model produce outputs relevant to conversations where the texter may be less likely to respond to the texter survey. Further, the use of multi-task learning enables the model to learn relationship between co-variates implicitly and therefore trains survey predictions implicitly through training auxiliary conversations on co-variate labels. For example, if there is a relationship between responses to a certain survey question and suicide risk, training the model to predict suicide risk on non-survey conversations will improve the model's ability to generalise predictions on that survey question. \\
\chapter{Model Implementation}
\label{chapter:implementation}

In this chapter, I will discuss the implementation of the Machine Learning models used in this project. I use increasingly complex models starting with two baselines, namely Random Forests using hand-crafted conversation features followed by TF-IDF word occurrence vectors. Next, I use a relatively small Residual Convolutional and Recurrent Neural Network (or CNN-RNN) with pre-trained word embeddings. I then turn to the use of BERT models, fine-tuning a pre-trained model on a Masked Language Modelling task. Due to Shout's conversations being generally longer than BERT's maximum sequence length of 512 tokens, I experiment with three potential methods of applying BERT longer sequences, namely Voting over BERT, which involves independently forming predictions on overlapping sub-sequences and combining the predictions from each sub-sequence; Attention Over BERT, which involves five methods of combining multiple BERT forward passes using an attention mechanism; and Transformer over BERT, a method slightly modified from \citet{hierarchicaltransformers} for end-to-end training. I also explore the benefit of multi-task training, by training on auxiliary labels including conversation metadata and non-target survey data. Results from these models are evaluated in the next chapter. \\

\section{Basic Features and TF-IDF}

Initial baseline experiments were used to assess problem difficulty, understand the importance of different types of features in predicting different labels, and justifying the need for more complex BERT models. The first method of document representation used is extraction of hand-crafted features. The features I use are based on conversations with individuals at Shout to understand what may be most useful for predicting labels such as demographics and conversation helpfulness, including the preliminary research in Chapter \ref{chapter:dataset}. These decided features are conversation duration, number of messages, words, and characters sent by each the CV and the texter, as well as the ratios of these values. Based on conversations with members of the Shout team, age or demographics might relate with the speed of responses, length of messages, or length of words. Similarly, length of conversation may vary by topic and may relate with how helpful texters find conversations. Relative number of messages might explain the relative involvement of each speaker, for example; the average length of words used by a speaker or the speed at which they type might represent their mental state; and the total length of a conversation might indicate level of distress.  Admittedly, this list is far from complete. With more time and domain expertise, proper feature engineering could be used to identify features from a larger set, such as time of day, time of year, the full distribution of time between messages, etc., or task-specific features such as the presence or frequency of certain key words. However, it is expected the later deep learning models will perform automatic feature engineering on message data, so these basic features will suffice as a baseline. \\

As a second baseline, I use TF-IDF vectors with the implementation in Scikit-learn \citep{scikit-learn}. I generate TF-IDF vectors separately for the texter, CV, and bot (i.e. the automated messages); I then concatenate these to form one vector per conversation. Bot messages are included in case they provide valuable metadata, for example about wait times. Tokenisation and stop-word removal are performed with library defaults and  words that occur in more than 40\% of conversations (domain-specific stop-words) or fewer than 1\% for CV and texter messages and 5\% for bot messages. These thresholds were chosen by manually scanning through term frequency to identify a reasonable boundary between stop-words and meaningful words; and similarly on lower end, the boundary aims to reduce very rare terms that increase likelihood of overfitting. I then perform Latent Semantic Analysis \cite{deerwester1990indexing} to reduce the dimensionality of the TF-IDF vectors, capturing redundancy in the vectors (e.g. if ``thank'' and ``you'' very frequently co-occur, this can be taken into account). LSA also makes these vectors more computationally manageable and reduces the ease of overfitting for labels with fewer examples. I use the Truncated SVD algorithm for dimensionality reduction from \citet{halko2009finding} for efficiency at the slight cost of being sub-optimal, due to the size and sparse nature of the TF-IDF vectors. The vector dimensionality after dimensionality reduction is 1000. \\

As a model, a Random Forest classifier is used for its ability to reduce model variance in addition to bias via its use of an ensemble of weak decision trees. Using a Random Forest also allows for complex relationships between words to be considered, rather than merely the presence or frequency of a single word. I introduce early stopping to the Random Forest, where the forest stops growing once a certain metric evaluated on a validation set plateaus, to further balance model bias and variance in training (See \verb|random_forest_early_stopping.py| for implementation). In terms of hyper-parameters, a balanced weighing of classes is used; a maximum tree depth of 4 for basic features and 5 for TF-IDF; and an increasing number of trees with a step size of five trees per epoch and a patience of 3 epochs before stopping training for the best validation metric. The Matthews Correlation Coefficient between the predicted and true label is used as the validation metrics, due to the imbalanced nature of many classes. \\

\section{CNN-RNN}

Next, I turn to training a deep CNN-RNN with pre-trained word embeddings. This model incorporates a number of elements from Deep NLP described in Chapter \ref{chapter:background}. The use of CNN layers allows for the detection of co-occurring words and phrases and the use of RNN layers allows for distant dependencies to be taken into account. The use of strided convolutions also helps reduce the size of the input to the RNN, thus improving efficiency. The RNN cell type used is the Gated Recurrent Unit (GRU); as explained in \ref{chapter:background}, GRU allows for nearly identical performance compared to LSTMs at a fraction of the computational cost.\\

To allow for comparison with BERT, the same tokenisation strategy is used for the CNN-RNN as for BERT and the embedding layer is reused from BERT. The use of pre-trained word embeddings allows the CNN-RNN to incorporate the meaning of words based on a much larger corpus. The CNN-RNN is expected to perform strictly worse than a BERT model; it has far fewer weights, has only one pre-trained layer, and does not use self-attention layers. However, it has the benefit of substantially quicker training time, and can therefore be useful in evaluating the impact of various factors on classification and understanding the difficulty of modelling different variables. \\

The CNN-RNN is designed as follows. First, a conversation is tokenized with special tokens added to indicate the start and end of a conversation and to separate between messages. An embedding layer, initialized with pre-trained word embeddings from BERT, then converts the tokens to corresponding embedding vectors. A pre-trained Byte-Pair Encoding tokenizer is used from RoBERTa which works by merging adjacent bytes to form increasingly large tokens \citep{sennrich2015neural}; this methodology also ensures that the dataset will have no unknown tokens, and will be able to correctly handle emojis. Tokens are also added to the tokenizer for common words that are not otherwise recognised like "NHS"; a [scrubbed] token for redacted words; for all bot messages that occur a great enough number of times; and a number of complete messages that occurs with sufficient frequency, such as initial and final messages sent by CVs. \\

In addition to word embeddings, for each token, a separate embedding is added corresponding with which speaker type wrote the token, similar to the learnt segment embedding in BERT \cite{bert}. This embedding mechanism is demonstrated in Figure \ref{fig:cnnrnn_embeddings}. \\

\begin{figure}[htb]
\setlength{\tabcolsep}{2pt}
\fontsize{8}{8}\selectfont
\begin{tabular}{cccccccccccc}
\multicolumn{1}{r}{\textbf{Input:}}            & {[}START{]}  & Help    & {[}SEP{]}  & Hi         & ,          & please     & tell       & me         & what's     & wrong      & {[}END{]}  \\
                                               &              &         &            &            &            &            &            &            &            &            &            \\\hline\\
\multicolumn{1}{r}{\textbf{Token Embeddings:}} & E\textsubscript{{[}START{]}} & E\textsubscript{Help}   & E\textsubscript{{[}SEP{]}} & E\textsubscript{Hi}        & E\textsubscript{,}         & E\textsubscript{please}    & E\textsubscript{tell}      & E\textsubscript{me}        & E\textsubscript{what's}    & E\textsubscript{wrong}     & E\textsubscript{{[}END{]}} \\\\
                                               & +            & +       & +          & +          & +          & +          & +          & +          & +          & +          & +          \\\\
\textbf{Speaker Embeddings:}                   & E\textsubscript{0}           & E\textsubscript{texter} & E\textsubscript{0}         & E\textsubscript{counselor} & E\textsubscript{counselor} & E\textsubscript{counselor} & E\textsubscript{counselor} & E\textsubscript{counselor} & E\textsubscript{counselor} & E\textsubscript{counselor} & E\textsubscript{0}        
\end{tabular}
\caption{Example of a conversation embedding.}
\label{fig:cnnrnn_embeddings}
\end{figure}

\begin{figure}[tb]
\centering
\includegraphics[width = 0.5\hsize]{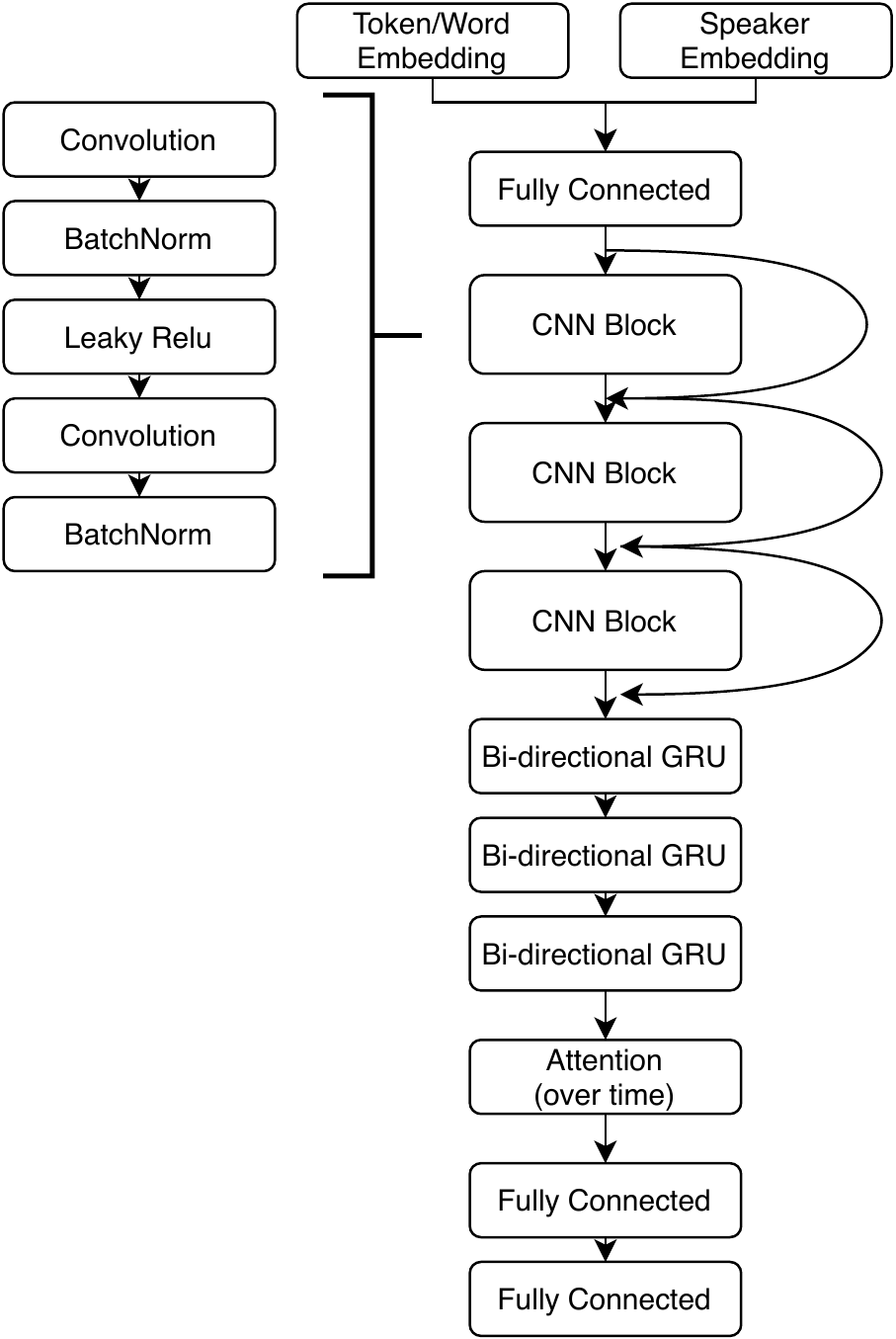}
\caption{Architecture of the CNN-RNN. Arrows denote skip connections.}
\label{fig:cnnrnn_architecture} 
\end{figure}

Embeddings are then passed through a fully connected layer. This is added to allow the model to convert embeddings into a more useful form for the CNN layers without needing to fine-tune all word embeddings; as some words might appear in the test set and not the training set, keeping word embeddings as close as possible to those that are pre-trained can reduce model overfitting. The model then applies three convolutional blocks, with residual `skip' connections, based on ResNet \citep{resnet}. The first block uses a stride of 1, the following two each use a stride of 2, and all convolutional layers use a kernel size of 3. \footnote{For skip connections about an underlying strided convolutional block, a single convolutional layer is applied with kernel size of 1 and the relevant stride.} Outputs from the CNN are then passed through three layers of bi-directional GRU to allow for modelling long-term relationship with a conversation. The six vector outputs from the GRU (i.e. one vector per layer per direction) are concatenated and passed through passed through two fully connected layers, separated by a LeakyReLU activation. A diagram of the CNN-RNN model is shown in Figure \ref{fig:cnnrnn_architecture}. \\

\subsection*{Multi-Task Training}

While Random Forests can only be trained to predict a single label at a time, neural networks can be trained to predict any number of labels simultaneously. I therefore experiment with multi-task training with the CNN-RNN to benefit from correlations between different labels. In the multi-task setting, the model is trained to output 489 outputs simultaneously; these include all questions from the CV post-conversation survey, the texter post-conversation survey; the CV skill labels, which will be discussed in \ref{chapter:measuring_skill}; and data on whether a conversation was escalated or led to an active rescue. Additionally, regular expression-based labels are added to extract ages from messages when a texter writes, e.g., "I'm 25", to push the model to pay attention to explicit statements of a texter's age. Merging this with texter survey data is likely unwise as this data is likely less reliably accurate than texter survey data; texters are more likely to be asked their age if they're young and might be less likely to tell the truth if they worry it'll affect how they're perceived. \\

Ordinal labels are converted into binary rank variables, such that an ordinal variable with labels 1, 2, 3 would be mapped to two binary variables with values (0,0), (0, 1), (1,1), as in \citet{ordinalloss}. Free text response labels are converted into word indicator vectors, with each position of the vector representing only whether a given word is present in the text. All words are stemmed, stop-words are removed, and only the most common words thereafter are kept. \\

Due to the sparsity of various labels, where many labels such as texter survey answers are absent for many conversations, it is important that only labels present for a given example are used in the loss function. Additionally the losses for each label are combined in a weighted way, ensuring that losses between different examples don't vary too much. To achieve this, first, the loss for all possible answers to a single question are combined by taking the mean; then, all questions that contain labels for a given example are combined by taking the mean. Binary Cross Entropy and Mean Squared Error are used as the loss functions for binary and continuous labels, respectively.

\section{BERT}

All remaining experiments involve use of BERT models. For all but the final experiment, I use a pre-trained ``DistilRoBERTa'' due to its smaller size and performance - experimenting with several models would be time consuming and unlikely to significantly change results. The somewhat bigger RoBERTa-base model is trained in the final experiments to understand the impact of increasing model size and for predictions As explained with regards to the CNN-RNN, the tokenizer is fine-tuned from that used to train the original model with domain-specific words and sequences, and learnt embeddings are added to indicate the speaker of each token. \\

I begin by fine-tuning the model on a Masked Language Modelling (MLM) task for five epochs with hyperparameters taken directly from \citet{bert} and \citet{roberta}. Sequences for the MLM task are 512 token sub-sequences extracted from conversations with 256 token overlap between sequences. A validation set is used to ensure that the model does not overfit to the training set in the MLM task. MLM training allows the model to capture semantic and syntactic information about conversations, accelerating learning and improving the model's ability to generalise in downstream tasks. Next, I fine-tune BERT in a supervised learning fashion, predicting the relevant set of labels. As with RoBERTa \citep{roberta}, there is no pre-train task associating the [CLS] tag at the beginning of sequences with the entire sequence, and therefore an attention layer is instead used to combine all BERT outputs, as explained in \ref{chapter:background}. \\

\begin{figure}[tb]
\centering
\includegraphics[width = 0.25\hsize]{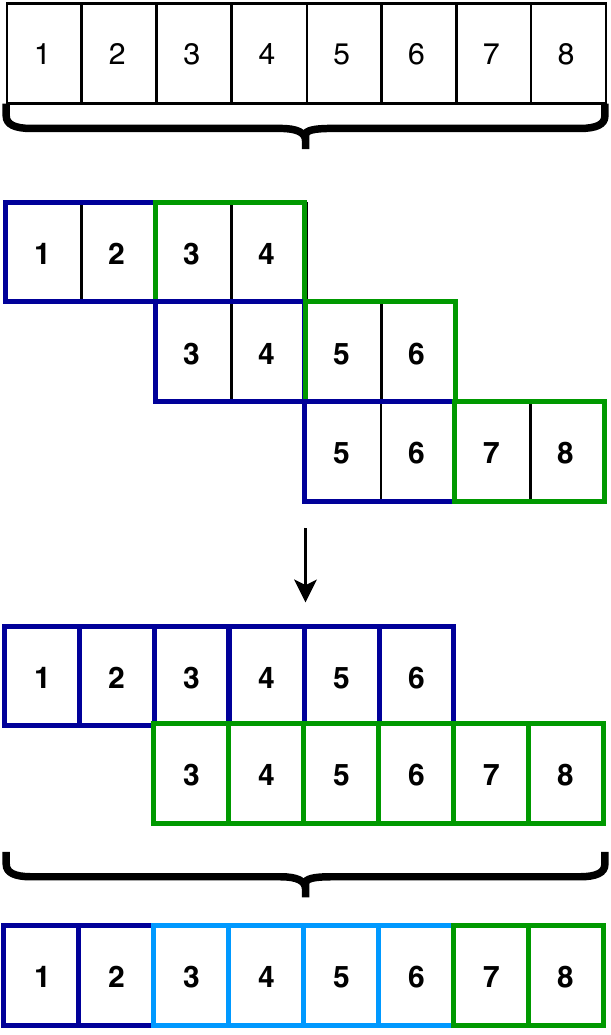}
\caption{A visualisation of how inputs are passed through BERT with overlap and then recombined. In this example, the maximum sequence length is 4 and an overlap of 2 is used between segments. The blue segments are the left-embeddings and the green segments are right-embeddings.}
\label{fig:bert_striding_method} 
\end{figure}

Using BERT with longer sequences than the maximum sequence length of 512 token is non-trivial. The method used here to overcome this is to combine outputs from multiple BERT forward passes over overlapping sequences with the maximum sequence length. Figure \ref{fig:bert_striding_method} shows how long sequences can be segmented with overlap, passed through BERT, and reshaped to form a full sequence embedding. I then experiment with three methods of increasing complexity for converting these BERT output embeddings into label predictions: voting, attention, and a transformer. \\

First, voting involves simply taking the mean of the model outputs from each sequence. Voting can be unweighted, where all sequences are taken account equally, or weighted, where sequence outputs are weighted less proportionally to the number of tokens that they contain that appear in another forward pass, reducing the effect of tokens that appear in two forward passes being counted twice. \\

Next, combining forward passes with the attention mechanism is slightly cleverer. At training time, BERT token embeddings can be combined using an attention layer, taking a feature-wise weighted mean based on the importance of each feature of each token. This attention mechanism can map any number of inputs to a single output and can therefore combine token embeddings from multiple forward passes. Again, to reduce the effect of overlapping tokens being counted twice, these token embeddings can be weighted less, this time by reducing the magnitude of the tokens' attention vectors. There are two ways to weigh down these attention vectors - before or after the softmax function is applied. Applying it beforehand can seem more logical because it allows softmax to normalize the attention scores to sum to 1, as is done in training; however, doing so can change the distribution of attention scores in an undesirable way. Weighing them down after softmax should affect the attention score distribution less, and can be followed by simple linear normalization to have attention scores add to 1. \\

Lastly, as discussed in the background (chapter \ref{chapter:background}), \citet{hierarchicaltransformers} find success combining BERT outputs with a small number of RNN layers or transformer layers. Due to the lack of full parallelisability, RNNs were prohibitively slow for experimentation in this project, but use of transformer layers proved useful. For my implementation of the Transformer over BERT or ToBERT model, slight modifications are made compared to the paper implementation to allow for end-to-end training. First, left- and right-embeddings of 256 token sub-sequences are concatenated feature-wise with 0 padding added for tokens with either no left or no right-embedding. These embeddings are combined with a two-layer neural network, passed through a two layer transformer, and combined to form a single embedding vector with an attention mechanism, and finally converted to a prediction with a linear layer. The two layer transformer inherits the same dropout, feature dimension, feed-forward dimension, and activation function from the underlying BERT model. Unlike BERT, the transformer model uses sinusoidal position embeddings to allow for unseen, or seldom seen, input lengths at test time as compared to training time.\\

As a note, all neural network models used here are implemented with the PyTorch library; the Weights and Biases platform \citep{wandb} is used for hyperparameter tuning, to monitor model training, and to compile model metrics; and the HuggingFace \citep{huggingface} library and repository is used for importing pre-trained BERT models.

\chapter{Model Evaluation}
\label{chapter:evaluation}

This chapter will evaluate model performances for each of the models described in chapter \ref{chapter:implementation} and will conclude with a discussion of future directions for model improvements. Results from each model across a small subset of labels can be found in Figure \ref{table:comparison_of_methods}, with the Matthews Correlation Coefficient (MCC) used as an evaluation metric for comparison of models due to the imbalanced nature of these labels. Unsurprisingly, increasingly complex models perform increasingly well with deep learning significantly outperforming baselines and BERT providing significant improvement upon the CNN-RNN. \\

% \begin{table}[htb]
% \fontsize{10}{12}\selectfont
% \begin{tabular}{lllllll}
% \textbf{} 
% &\textbf{\begin{tabular}[c]{@{}l@{}}
% Age: 21 \\ or under\end{tabular}} &
% \textbf{\begin{tabular}[c]{@{}l@{}}
% Heterosexual/\\ Straight\end{tabular}} & \textbf{\begin{tabular}[c]{@{}l@{}}
% Topic:\\ Self Harm\end{tabular}} & 
% \textbf{\begin{tabular}[c]{@{}l@{}}
% Topic:\\ Depression\end{tabular}} & \textbf{\begin{tabular}[c]{@{}l@{}}Texter:\\ Helpful\end{tabular}} & \textbf{\begin{tabular}[c]{@{}l@{}}
% CV: Suicidal \\ Capability\end{tabular}} \\
% \textbf{Features} & 0.180 & 0.079 & 0.116 & 0.115 & 0.241 & 0.349 \\
% \textbf{TF-IDF} & 0.455 & 0.158 & 0.581 & 0.319 & 0.286 & 0.531 \\
% \textbf{ResNet CNN RNN} & 0.518 & 0.168 & 0.738 & 0.264 & 0.428 & 0.739 \\
% \textbf{Voting Over BERT*} & 0.723 & \textbf{0.211} & 0.747 & 0.454 & 0.480 & 0.741\\
% \textbf{Attention Over BERT*} & 0.737 & 0.199 & 0.755 & 0.454 & 0.505 & 0.753\\
% \textbf{Transformer over BERT*} & \textbf{0.751} & 0.206 & \textbf{0.769} & \textbf{0.482} & \textbf{0.537} & \textbf{0.783}\\ 
% \end{tabular}
% \caption{Comparison of results from each of the main models across a select subset of labels. Metric shown is the Matthews Correlation Coefficient. For Voting over BERT, the weighted method is used; and for Attention over BERT, the weight before softmax method is used. The * models all reuse the same pre-trained distilRoBERTa model}
% \label{table:comparison_of_methods}
% \end{table}

\begin{table}[htb]
\centering
\fontsize{10}{12}\selectfont

\begin{tabular}{lccccccc}
\textbf{} & \textbf{\begin{tabular}[c]{@{}c@{}}Basic\\ Features\end{tabular}} & \textbf{TF-IDF} & \textbf{CNN-RNN} & \textbf{\begin{tabular}[c]{@{}c@{}}BERT*\\ Voting\end{tabular}} & \textbf{\begin{tabular}[c]{@{}c@{}}BERT*\\ Attention\end{tabular}} & \textbf{ToBERT*} & \textbf{ToBERT-base} \\
\multicolumn{8}{l}{\textbf{Topic:}} \\
Self Harm & 0.116 & 0.581 & 0.738 & 0.747 & 0.755 & 0.769 & \textbf{0.777} \\
Depression & 0.115 & 0.319 & 0.264 & 0.454 & 0.454 & \textbf{0.482} & 0.480 \\
Substance Use & 0.014 & 0.3477 & 0.068 & 0.5416 & 0.555 & 0.555 & \textbf{0.559} \\
\textbf{Suicide Risk:} &  &  &  &  &  &  &  \\
Desire & 0.255 & 0.521 & 0.733 & 0.068 & 0.743 & 0.752 & \textbf{0.758} \\
Capability & 0.349 & 0.531 & 0.739 & 0.741 & 0.753 & 0.783 & \textbf{0.793} \\
\multicolumn{8}{l}{\textbf{Texter Survey:}} \\
Helpful? & 0.241 & 0.286 & 0.428 & 0.480 & 0.505 & \textbf{0.537} & 0.531 \\
Age $\leq$ 21 & 0.180 & 0.455 & 0.518 & 0.723 & 0.737 & 0.751 & \textbf{0.761} \\
Heterosexual & 0.079 & 0.158 & 0.168 & \textbf{0.211} & 0.199 & 0.206 & 0.204 \\
Race: White & 0.158 & 0.080 & 0.079 & 0.166 & 0.182 & \textbf{0.219} & 0.172 \\
Gender: Male & 0.05387 & 0.2491 & 0.142 & 0.307 & 0.304 & \textbf{0.384} & \textbf{0.384}
\end{tabular}

\caption{Comparison of results from each of the main models across a select subset of labels. Metric shown is the Matthews Correlation Coefficient. For Voting over BERT, the weighted method is used; and for Attention over BERT, the weight before softmax method is used. The * models all reuse the same pre-trained distilRoBERTa model and ToBERT-base uses a pretrained RoBERTa-base model}

\label{table:comparison_of_methods}
\end{table}

\section{Baseline Models}

The variation in performance between the baseline models (Basic Features, TF-IDF and CNN RNN) varies significantly between labels, which can be for a number of reasons. The basic features model succeeds when a label correlates well with length- or speed-related features. For example, the basic features model performs relatively well in predict texter helpfulness reviews, perhaps because longer conversations are more helpful, conversations with faster responses are more helpful, or a something of the like. Predictions of suicidal capability are also predictable from these features, though the relationship is very unlikely causal. It is more likely that conversations with high suicide risk are escalated and therefore lead to longer conversations, ones with faster response time, or other factors that are not \textit{caused} by suicidal capability. \\ 

TF-IDF performs somewhat surprisingly well at predicting age, as compared with the CNN-RNN, and performs even better at predicting the topic of depression. TF-IDF can be expected to perform well on detecting topics as topics can typically be identified by the presence and frequency of specific words and phrases. For example, the topic "self-harm" is predicted relatively well by TF-IDF, likely because conversation regarding self-harm often use specific words like "cutting." Word usage may also play a large role in differentiating between individuals of different ages, as discussed in Chapter \ref{chapter:dataset}. By contrast, the CNN-RNN has the benefit of having access to long-term relationships between tokens, which may be more helpful for some labels than others. \\

\section{Multi-Task Learning}
\label{section:eval_multitask}
The TF-IDF model's superior performance over the CNN-RNN suggests that Multi-Task Learning may have a negative effect in the ability to model certain labels. Though the good TF-IDF performance the depression label may be explainable due to the types of features relevant for detecting topics, the CNN-RNN has access to exactly the same features. The likely reason for TF-IDF's superior performance on the depression label is that a new Random Forest is trained independently on each label; as topics are labelled in many conversations, multi-task learning may be less necessary and less helpful for identifying the presence of some topics.\\

While multi-task learning may be beneficial for helping a model identify the most important features and to generalize better, it makes the sacrifice of using only a small subset of model weights on each label. Multi-task learning may also lead to the model making compromises during optimisation - paying attention to some labels, especially those that are easier to predict, than others. It is likely that the CNN-RNN model is too small to maintain relevant features for predicting all labels, and therefore will have inferior performance on some labels over others. There are at least two solutions to this problem. First, a much larger model can be used - and this is likely the correct path to take for the best possible results. The T5 model \citep{t5} discussed in Chapter \ref{chapter:background} uses about 11 billion parameters, which undoubtedly is helpful for it's multi-task learning. However, even the authors there discuss the negative impact on T5 from multi-task only training due to task interference. Alternatively, a model can be pre-trained with a multi-task objective and then fine tuned on one task or a small set of tasks. \\

To evaluate the improvement of fine-tuning with relatively small model, the CNN-RNN was re-trained on two labels separately, once from scratch, and once from a multi-task training checkpoint. These results can be found in Table \ref{table:cnnrnn-multitask} on the texter survey helpfulness question and the suicidal timeframe question. It can be seen that fine-tuning slightly improved results for both questions, compared to training it alone or only performing multi-task training. Interestingly, the helpfulness question performed slightly better trained alone, compared to only being training in the multi-task setting, while the timeframe label performed slightly worse alone. \\

\begin{table}[hb]
\begin{tabular}{lll}
\textbf{} & 
\textbf{\begin{tabular}[c]{@{}l@{}}Texter Survey: \\ Helpful\end{tabular}} &
\textbf{\begin{tabular}[c]{@{}l@{}}Risk Ladder: \\ Timeframe\end{tabular}} \\
\textbf{All Tasks} & 0.4282 & 0.6989 \\
\textbf{Single Task} & 0.4475 & 0.6865 \\
\textbf{Fine-tuned} & 0.4667 & 0.7113 
\end{tabular}
\caption{Comparing single-task, multi-task, and multi-task fine tuning using the CNN-RNN model on two specific labels}
\label{table:cnnrnn-multitask}
\end{table}

A likely explanation of this is that the texter survey helpfulness question is given less attention by the optimisation due to its relatively small contribution to the loss function, being only one of a large number of labels and also only being present in a relatively small subset of conversations. As a result, the intermediate feature vectors contain less signal for predicting helpfulness. That said, the label clearly benefits from multi-task training, as seen in the improved metric in the fine-tuned setting. By contrast, the timeframe question label is present in nearly all conversations and therefore the benefit from multi-task training leads to a slightly better metric, even in the multi-task-only setting. \\

Importantly, the difference between metrics in all three cases is not very significant. While fine-tuning provides benefit, it also requires training separate models for each label and passing through them all in producing predictions. Compromises can therefore be made, for example fine-tuning on several related labels simultaneously; fine-tuning only the final layers of the model and not the entire model, to reuse certain computations; or simply using the multi-task model where its results are sufficiently strong to outweigh the benefit from fine-tuning.\\

\section{BERT}

The use of BERT significantly improves performance across all labels, which is a mainly a testament to the problem difficulty. As discussed in the background, BERT's transformer architecture is able to model long term relationships far better than CNNs or RNNs, and BERT's semi-supervised approach allows for better generalisation. \\

\begin{figure}[htb]
\centering
\includegraphics[width = 0.6\hsize]{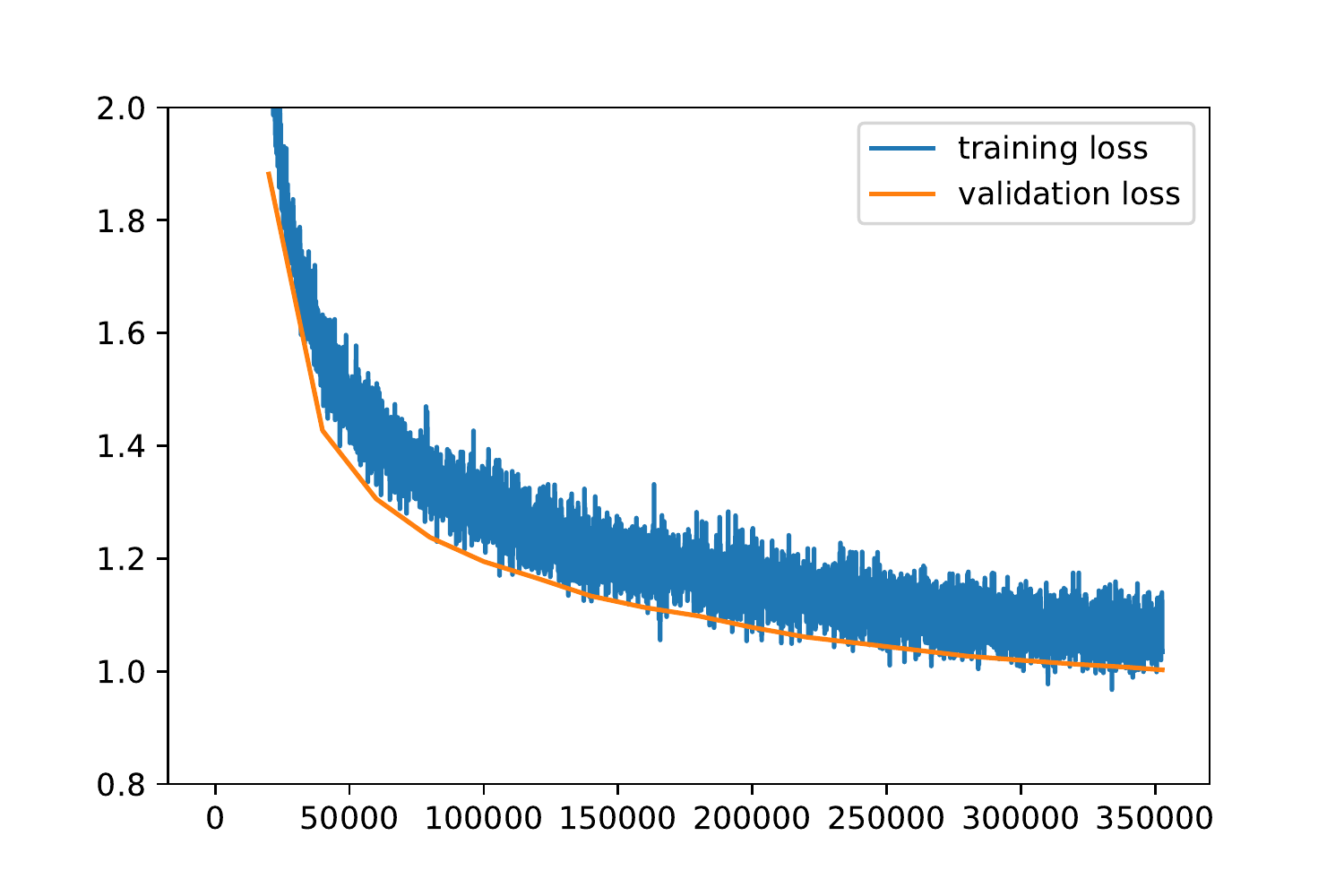}
\caption{Training and Validation loss curves for Masked Language Model training (distilRoBERTa)}
\label{fig:mlm_loss} 
\end{figure}

As explained in the previous chapter, BERT was first fine-tuned with a Masked Language Modelling task. The loss curves on a training and hidden validation set, shown in figure \ref{fig:mlm_loss}, show a smooth loss and a good balance between bias and variance. BERT can also be manually evaluated by selecting random conversation segments from the test set, masking random tokens, and generating predictions. As expected, outputs were often correct and matched the tokens that a human likely predicted. Interestingly, however, there were several instances where BERT predicted correctly tokens that would have been very difficult for a human to predict. On one occasion, in a conversation about addiction, BERT correctly predicted the full phrase ``12-step program'' which did not appear anywhere else in the conversation. A few times, the model incorrectly identified noise where a typo was made by the texter, but actually replaced the typo with the correct spelling and grammar. And finally, a few experiments were performed where the speaker tokens for a message were swapped in BERT's inputs, while the underlying message text remained intact. Interestingly, a degree of style transfer was observed, especially with messages originally from texters made to sound more professional, while the content remained mostly intact. This exploration helps justify proceeding with the BERT model and provides an understanding that BERT was able to successfully model conversations. \\

Returning to supervised learning, as discussed in chapter \ref{chapter:implementation}, three methods are used for applying BERT to conversations that exceed BERT's maximum feasible sequence length - voting, attention, and a transformer. Unsurprisingly, the transformer model's performance significantly exceeds that of voting and attention, most likely due to its ability to track long-term relationships across a conversation. For labels where long-term relationships are less relevant, such as demographic labels, such methods are likely to provide less benefit compared to using a larger model with comparable training time. Attention over BERT proves to be an effective method for combining BERT outputs, providing improvement over voting on a number of labels with virtually no change in the BERT model or training loop. Interestingly, for the Heterosexual/Straight label, voting outperforms Attention or Transformer over BERT. This may simply be an artefact of multi-task learning, where more difficult to predict labels may receive less training, and might disappear with fine-tuning or with a much larger model. \\

\begin{table}[htb]
\centering
\setlength{\tabcolsep}{2pt}
\fontsize{10}{12}\selectfont
\begin{tabular}{lllllll}

 &
  \textbf{\begin{tabular}[c]{@{}l@{}}Age: 21\\ or under\end{tabular}} &
  \textbf{\begin{tabular}[c]{@{}l@{}}Heterosexual/\\ Straight\end{tabular}} &
  \textbf{\begin{tabular}[c]{@{}l@{}}Topic: \\ Self Harm\end{tabular}} &
  \textbf{\begin{tabular}[c]{@{}l@{}}Topic: \\ Depression\end{tabular}} &
  \textbf{\begin{tabular}[c]{@{}l@{}}Texter: \\ Helpful\end{tabular}} &
  \textbf{\begin{tabular}[c]{@{}l@{}}CV: Suicidal \\ Capability\end{tabular}} \\
\textbf{\begin{tabular}[c]{@{}l@{}}Voting \\ Unweighted\end{tabular}} &
  0.723 &
  0.204 &
  0.748 &
  0.453 &
  0.486 &
  0.737 \\
\textbf{\begin{tabular}[c]{@{}l@{}}Voting \\ Weighted\end{tabular}} &
  0.723 &
  \textbf{0.211} &
  0.747 &
  0.454 &
  0.480 &
  0.741 \\ \hline
\textbf{\begin{tabular}[c]{@{}l@{}}Attention\\Unweighted\,\end{tabular}} &
  0.711 &
  0.210 &
  0.761 &
  0.449 &
  0.486 &
  \textbf{0.764} \\
\textbf{\begin{tabular}[c]{@{}l@{}}Weight after \\ Softmax\end{tabular}} &
  \textbf{0.737} &
  0.199 &
  0.755 &
  \textbf{0.454} &
  \textbf{0.505} &
  0.753 \\
\textbf{\begin{tabular}[c]{@{}l@{}}Weight before \\ Softmax\end{tabular}} &
  0.731 &
  0.210 &
  \textbf{0.766} &
  0.452 &
  0.480 &
  0.760
\end{tabular}\caption{Comparison over methods for combining BERT forward passes}
\label{table:attention_over_bert}
\end{table}

Voting and Attention over BERT methods are compared in Table \ref{table:attention_over_bert}. As explained in chapter \ref{chapter:implementation}, segments passed through BERT contain overlap, to ensure that all adjacent context can be taken into account by BERT. This can lead to over-consideration of segments that appear twice, which can be rectified by weighting less these segments. For voting, the weighted and unweighted variations had very similar results, with the weighted method slightly outperforming the unweighted variation, but within a reasonable margin of error. For Attention over BERT, the weight after softmax method generally outperforms the alternatives, but not very significantly. As predicted, Attention over BERT outperforms the simpler voting method. \\

Finally, the end-to-end trained ToBERT model outperforms the other two BERT methods on nearly all labels. This is expected, because unlikely ordinary BERT, ToBERT has access to the entire conversation at once and can therefore derive signal from very long term dependencies. Additionally, ToBERT has a greater number of model weights, due to the two additional layers. 

\begin{table}[htb]
\fontsize{10}{12}\selectfont
\centering
\begin{tabular}{llllll}
\textbf{} & \multicolumn{1}{c}{\textbf{MCC}} & \multicolumn{1}{c}{\textbf{Avg Precision}} & \multicolumn{1}{c}{\textbf{F1 Score}} & \multicolumn{1}{c}{\textbf{AUC-ROC}} & \multicolumn{1}{c}{\textbf{Accuracy}} \\
\textbf{Topic:} & \multicolumn{1}{c}{} & \multicolumn{1}{c}{} & \multicolumn{1}{c}{} & \multicolumn{1}{c}{} & \multicolumn{1}{c}{} \\
Self Harm & 0.777 & 0.894 & 0.815 & 0.974 & 0.938 \\
Depression & 0.480 & 0.741 & 0.662 & 0.829 & 0.762 \\
Substance Use & 0.559 & 0.611 & 0.570 & 0.960 & 0.978 \\
\textbf{Suicide Risk:} & \multicolumn{1}{c}{} & \multicolumn{1}{c}{} & \multicolumn{1}{c}{} & \multicolumn{1}{c}{} & \multicolumn{1}{c}{} \\
Desire & 0.758 & 0.895 & 0.838 & 0.956 & 0.893 \\
Capability & 0.793 & 0.875 & 0.822 & 0.981 & 0.949 \\
\textbf{Texter Survey:} & \multicolumn{1}{c}{\textbf{}} & \multicolumn{1}{c}{\textbf{}} & \multicolumn{1}{c}{\textbf{}} & \multicolumn{1}{c}{\textbf{}} & \multicolumn{1}{c}{\textbf{}} \\
Helpful? & 0.531 & 0.985 & 0.943 & 0.908 & 0.900 \\
Age > 21& 0.761 & 0.943 & 0.858 & 0.957 & 0.884 \\
Heterosexual & 0.204 & 0.709 & 0.658 & 0.652 & 0.610 \\
Race: White & 0.172 & 0.943 & 0.922 & 0.673 & 0.859 \\
Gender: Male & 0.384 & 0.443 & 0.463 & 0.803 & 0.863
\end{tabular}
\caption{Results from the final ToBERT-base model, i.e. Transformer over RoBERT-Base. A greater selection of metrics are shown to better understand model performance.}
\label{table:all_metrics}
\end{table}

\section{Irreducible Error}
The last model shown in \ref{table:comparison_of_methods} is ToBERT-base, a ToBERT model built upon a pre-trained RoBERTa-base model rather than distilRoBERTa-base. A greater number of metrics for the model can be found in \ref{table:all_metrics}. Using a slightly larger model allows for a better understanding of the degree to which larger models will improve performance on each of the relevant tasks. It is worth noting that the difference between the two underlying BERT models used here is the number of transformer layers used rather than the embedding dimension between layers (768), so it is unlikely to reduce label interference effects discussed in the above Section \ref{section:eval_multitask}.

Interestingly, ToBERT-base does not universally outperform ToBERT, though it does outperform it most of the time. It is unlikely that this is due to model overfitting as the ToBERT models are trained for just three epochs. A better explanation for this is the presence of \textit{irreducible error}. By irreducible error, I mean intrinsic label noise that cannot be modelled in terms of the relevant inputs, no matter how powerful the model is. For example, the maximum possible degree of accuracy in predicting race of sexual orientation on the basis of a conversation may not be very high. For some of these labels, increasing model complexity and especially increase of model size as in the two ToBERT models does little to improve performance, a strong indication of irreducible error.\\

Another source of irreducible error is mislabelling, which can limit the maximum metrics achievable on the given test set. For example, depression is only correctly identified in a mere 76.2\% of test conversations, a task likely far easier than predicting texter helpfulness reviews, which are correctly predicted correctly in 90.0\% of test texter survey conversations. The most likely explanation for this failure to identify depression in conversations is label noise due to mislabelling.If this is the case, it is likely that reducing label noise, for example by evaluating on gold standard labels, may indicate superior model performance to that indicated by the test set. A proper analysis of noise from mislabelling labels will be addressed specifically for the Suicide Risk Ladder in Chapter \ref{chapter:suicide_risk_ladder}. \\

\section{Discussion}

Evaluation of increasingly complex models has justified the need for a large model of the size and complexity of BERT. It has also shown the benefit of semi-supervised learning through pre-training and the benefit of multi-task learning. Fine-tuning on specific labels after multi-task training has been shown to improve model performance, especially in relatively small models. Going forward, one can expect that a significantly larger model may provide significantly better performance, and that fine-tuning on labels will improve performance, especially on harder to predict or less frequently occurring labels. However, fine-tuning on many different labels would require producing a large number of models, drastically increasing memory requirements and time for training and inference. For this reason, the multi-task model may be treated as ``good enough'', even though improvements are surely possible. In future research, in addition to using a bigger model, compromises may be made between separately using a multi-task only model or a number of fine-tuned models. For example, early model layers can be frozen after multi-task training and later layers can be differentiated by task. Or alternatively, labels can be fine-tuned in groups that are expected to correspond with very similar features. \\

As discussed in \ref{chapter:implementation}, the Transformer over BERT model adjusts the methodology from \citet{hierarchicaltransformers}, training the transformer and BERT simultaneously, end-to-end. The benefits of training end-to-end were not evaluated here, but it was found that training end-to-end caused little additional computational time. For the Transformer over BERT model, the large transformer acting on the BERT outputs dominates training time due to transformers' $O(N^2)$ computational time, with $N$ representing sequence length. As a result, the cost of training end-to-end compared to training BERT first and then the transformer was negligible. In future research, the difference in performance may be evaluated. Another methodology considered and not experimented with would be the reducing sequence length between transformers, for example with Maximum-Pooling layer, to significantly reduce training time. This was not necessary here simply because the maximum sequence length was not excessively large.\\

\chapter{Participation Bias}
\label{chapter:participation_bias}
Participation bias can be a significant issue in using survey data to analyse a service, as discussed in Chapter \ref{chapter:background}. A few examples are provided there for gender participation bias \citep{chang2009national, hill2013wikipedia}, where gender may correlate with the likelihood of filling out a certain voluntary survey; as well as polarizing participation bias \citep{chamberlain2017give}, where individuals who fill out surveys regarding a service are more likely to have had very positive or very negative experience of a service. Consider texter survey answers to the question "Did you find this conversation helpful?" - it is valuable to understand whether texters who answer the question in the survey are more likely to respond to the affirmative or to the negative compared to texters who don't respond. The studies mentioned above just a few of many handling participation bias, but the research into the topic can hardly be used to assume any specific bias in the Shout texter survey. Because Shout's texter survey is with individuals in mental health crisis, because the texter survey is sent over text message, and for a number of other reasons, Shout's texter survey participation bias is likely unique and cannot be understood by simply generalising from research into other surveys. \\

A major objective of this project is to assess the presence of survey participation bias, estimate its extent, and reverse it. As discussed in Chapter \ref{chapter:background}, participation bias is typically analysed by comparing the distribution of the survey sample with the corresponding population. This can be done by either using existing knowledge of the population (e.g. the UK has a roughly a 50/50 gender split) or using an additional survey with very high response rate. Neither method is possible with the Shout dataset as the population in question is Shout texters about whom very little is known; in fact, learning about that population is the very goal of this reserach. Additionally, a high response rate follow-up survey is simply not possible with Shout's emphasis on anonymity. \\ 

Deep learning provides a methodology for assessing the presence and degree of participation bias. Given the final model developed in the previous chapter which is able to predict, with reasonable accuracy, how a texter will answer this question, participation bias can be assessed by comparing model predictions on the set of conversations where the texter responds to the texter survey and the set of conversations where the texter does not. If there is no participation bias, we should expect these distributions to be identical; otherwise, we should expect to find a difference in the distributions. Beyond testing for the presence participation bias, it is valuable to understand its magnitude and to reverse it, to the extent possible. With a perfect Machine Learning model, modelling participation bias is trivial, amounting to simply predicting responses and using them as a ground truth label. However, where a model is imperfect and suffers from noisy predictions, reversing participation bias is still possible, as will be shown; in fact, this project will attempt to use an imperfect Machine Learning classifier to generalise from the texter survey and learn about the population of Shout texters. To do so, it will be necessary to develop a mathematical model for model noise, estimating the relationship between average model predictions over a group and the ground truth average for that group. In this chapter, I will develop such a model specifically for predicting the true ratio of a binary variable in a group given noisy model predictions for that variable in that group. \\

As with any mathematical model, simplifying assumptions must be made, limiting the potential use of results, and such assumptions and limitations must be made clear. Further, this model is intended mainly as a proof of concept of the impact of this project, and to produce quantifiable results for use in understanding Shout's texters and improving Shout's service. Future improvements of this model will increase its general applicability. Due to the simplifying assumptions, empirical research prove helpful for demonstrate and assessing the mathematical model's power and limitation. To this end, I've developed and trained toy models on toy dataset, with code made publicly available, to assess the abilities of this mathematical model. As will be made clear, despite the model's limitations, it nonetheless performs very well.\\

\section{Mathematical Model}
For simplicity, we will consider here a Machine Learning model trained on binary classification problem. Let us assume that before training, the relevant dataset is split into three parts - a training set, a validation set, and an evaluation set. The training and validations sets each have a mean (i.e. $\frac{\text{ Positives}}{\text{Positives} + \text{ Negatives}}$) very close to $p_{train}$ and the evaluation set has a mean of $\mu$, with $p_{train}$ not necessarily equal to $\mu$. Our goal is to understand whether $\mu = p_{train}$ or more generally, to learn the value $\mu$. The model is first trained as much as possible on the training set without overfitting. Next, it is calibrated on the training set, by which I mean that a decision boundary is chosen on the model outputs such that the mean of the predictions after calibration is the same as the mean of the true labels. Next, it is tested on the validation set to obtain performance metrics, most importantly the Matthews Correlation Coefficient (MCC) between the true and predicted labels. Finally, the model produces predictions on the evaluation set, using the decision boundary from calibration. Denote the prediction mean on the evaluation set $\hat \mu$ \\

Let $Y$ be the Bernoulli distributed Random Variable for the label of a single example in the evaluation set and $\Yhat$ refer to the Random Variable the model's predicted label for that example. Let $m$ be a measure of model accuracy, which will empirically be shown to be closely related to the Matthews Correlation Coefficient (MCC); conversely, $(1-m)$ may be called the degree of noise. The model predictions are dependant on the true labels but will only be correct, on average, $m$ of the time. The other $(1-m)$ of the time, the model will output from the noise distribution. Because the model is trained on a specific training set, the expectation of the noise distribution is the same as the expectation of the training set, i.e. $p_{train}$.

\begin{equation}
Y = \B(\mu), \quad
\Yhat = \begin{cases} 
      y & m \\
      \B(p_t) & 1 - m
   \end{cases}
\end{equation}

For $m=0$, the model's guesses are completely random and simply extracted from the noise distribution $\B(p_t)$. This corresponds with a model that has not been trained at all, but merely calibrated. Conversely, a perfect model will have $m=1$ and will only ever output correct predictions, i.e. $Y=\Yhat$. For $p_t=\mu$, where the training and evaluation set are identically distributed, the noise will be unbiased; while if $p_t\neq \mu$, the noise will be biased by the training set mean, as is expected. \\

\begin{figure}%[tb]
\centering
\includegraphics[width = 0.5\hsize]{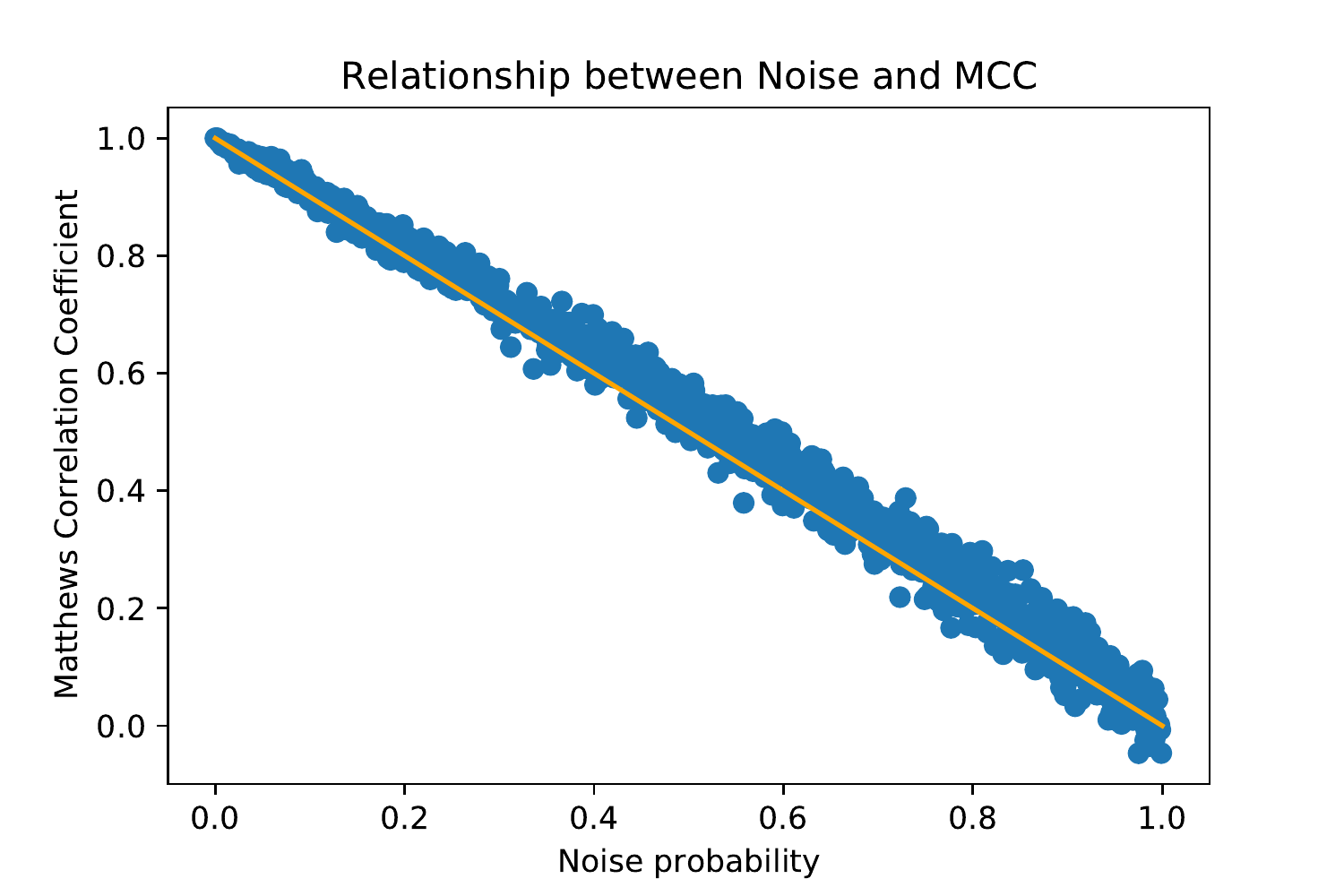}
\caption{Demonstration of the linear relationship between MCC and noise probability in noise model, using 1000 simulations with $p_{train}=p_{eval}=0.5$, each with N=10000.}
\label{fig:noise_mcc}
\end{figure}

The relationship between the degree of noise and the MCC can be understood intuitively or empirically. The relationship between $\Yhat$ and $Y$ is present $m$ of the time, so $m$ of the variation in $\Yhat$ is due to $Y$ and the rest is due to randomness. Empirically, one can find this relationship through simulations. In Figure \ref{fig:noise_mcc}, numerical simulations are performed showing the relationship between the noise parameter and the empirical MCC on a validation set of size 10000.\\

\subsection*{Modelling Predictions}

For testing the hypothesis that participation bias is not significant, we can use the above model to determine the likelihood of a certain predicted mean of an evaluation set ($\hat\mu$), given the true mean ($\mu$). For a single example, the conditional probability distribution is as follows:

\begin{align}
    \PP(\Yhat = 0 | Y = 0) =& m + (1-m)(1-p_t) \\
    \PP(\Yhat = 1 | Y = 0) =& (1-m) p_t \\
    \PP(\Yhat = 0 | Y = 1) =& (1-m)(1-p_t) \\
    \PP(\Yhat = 1 | Y = 1) =& m + (1-m)(p_t)\\
    \PP(Y = 0|\mu) =& 1-\mu \label{eq:py0}\\
    \PP(Y = 1|\mu) =& \mu 
\end{align}

Extending to the case of multiple examples, the likelihood may be understood in terms of $h$ positive (``heads'') and $t$ negative predictions (``tails''), given a true evaluation set mean of $\mu$. Let $i$ of the $h$ positives and $j$ of the $t$ negatives be correct, while the other $h-i$ and $t-j$ will be based on the noise distribution. 

\begin{align}
    \PP(h,t|\mu) =& {h+t \choose h} \sum_{i=0}^h \sum_{j=0}^t {h \choose i}{t \choose j} m^{i+j} (1-m)^{t+h-i-j} \mu^i (1-\mu)^j p_t^{h-i} (1-p_t)^{t-j} \\
    =& {h+t \choose h} \sum_{i=0}^h {h \choose i} (m\mu)^i ((1-m)p_t)^{h-i}  
    \sum_{j=0}^t {t \choose j} (m(1-\mu))^j ((1-p_t)(1-m))^{t-j}  \\
    =&  {h+t \choose h} \left(m\mu + (1-m)p_t)\right)^h \left(m(1-\mu) + (1-p_t)(1-m)\right)^t
    % \label{eq:likelihood}
\end{align}

This distribution clearly does not lend itself well to developing a confidence interval over the likelihood. Fortunately, we're interested in $\hat\mu$, which is a sample mean, and hence governed by the Central Limit Theorem; it should thus suffice to obtain only the expectation and variance for $\PP(\hat\mu|\mu)$, as long as the evaluation set is sufficiently large. Further, because each $y_i$ is independent and identically distributed, the expectation and variances are linear, and it should suffice to obtain the expectation and variances of $\PP(\Yhat|Y)$. The expectation is simply:

\begin{equation}
    \E[\Yhat|Y=y] = \PP(\Yhat = 1 | Y=y)  = m y + (1-m) p_{train}
\label{eq:pypred1}
\end{equation}

For obtaining the variance, we can use the fact that $\Yhat\in \{0,1\}$:
\begin{align}
\Var[\Yhat|Y=y] =& \E[\Yhat^2|Y=y] - \E[\Yhat|Y=y]^2\\ 
=& \E[\Yhat|Y=y] - \E[\Yhat|Y=y]^2 \\
=& (m y + (1-m) p_{train})(1 - m y - (1-m) p_{train})
\end{align}

Because each prediction-label pair in the evaluation set is independent and identically distributed from any other, the expectation and variance of a sum of a number of these predictions will simply be the sum of each expectation and variance; i.e.

\begin{align}
\E[\hat\mu|\mu] =& \frac{1}{n}\sum_{i=1}^n{\E[\yhat_i|y_i]} = m\mu + (1-m)p_{train}
\label{eq:indp_linearity_expectation}\\
\Var[\hat\mu|\mu] =&\frac{1}{n}\sum_{i=1}^n{\Var[\yhat_i|y_i]} = \mu\Var[\Yhat|Y=1] + (1-\mu) \Var[\Yhat|Y=0]\\
    \end{align}

The standard error of the mean coincides with that from the Central Limit Theorem - simply:

$$\sqrt{\Var[\Yhat|Y]/n}$$

\subsection*{Empirical Justification}

\begin{figure}%[tb]
\centering
\includegraphics[width = 0.5\hsize]{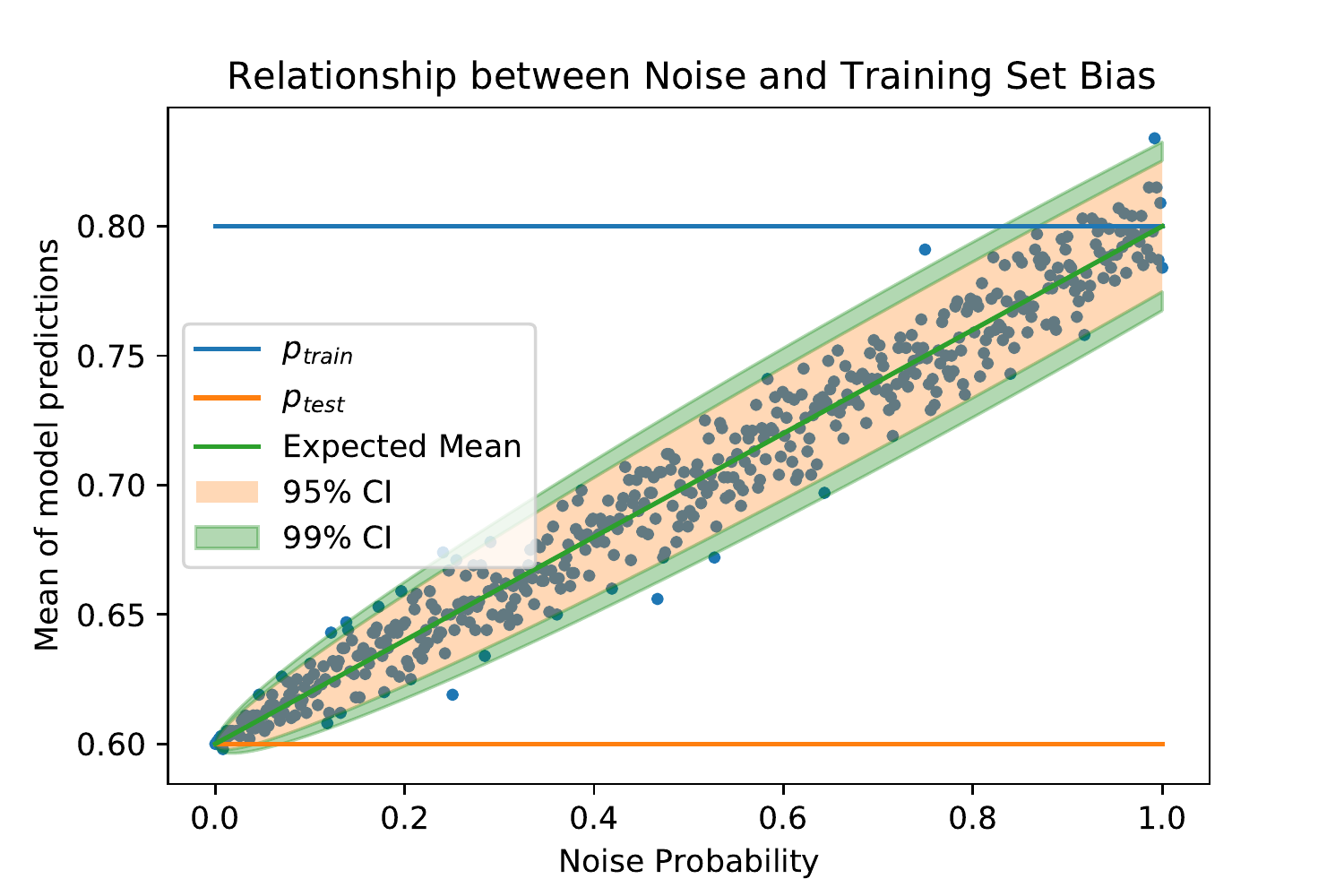}
\caption{Demonstration of the relationship between noise and bias towards the training set mean, using 1000 simulations each with $N=1000$ with $p_{train}$=0.8, $p_{eval}$=0.6}
\label{fig:noise_bias}
\end{figure}

These results can again be empirically verified. Figure \ref{fig:noise_bias} plots results from numerical simulations under the noise model for evaluation sample mean for a fixed sample size $N=1000$ and varying degrees of noise. As can be seen, the predicted means vary linearly from the true evaluation set mean to the training set mean as noise is linearly added. Confidence intervals based on the variance also seem correct, justifying the modelling of variance. \\

Importantly, the aim of this model is applicability to reversing bias with imperfect Machine Learning models. To fully justify this mathematical model, I've run simulations using a toy Convolutional Neural Network (CNN) on the MNIST dataset of handwritten digits \citep{mnist}. The CNN is trained on the binary task of predicting whether the number it receives is a five or not with a Binary Cross-Entropy loss function. Examples are intentionally removed from the training set to artificially change the proportions; as a result, 24.77\% of the training set are fives while 8.92\% of the test set are fives. The toy model is naively designed by simply alternating convolutional and ReLU layers three times, with a total of 6,073 weights; it is trained for one epoch only with a batch size of 64. A hidden validation set with proportions identical to the training set is used to evaluate the model's MCC. Again, calibration involves choosing a decision boundary to split the training set according to its correct proportions, and the MCC is calculated with this calibration. \\

\begin{figure}[htb]
\centering
\includegraphics[width = 0.5\hsize]{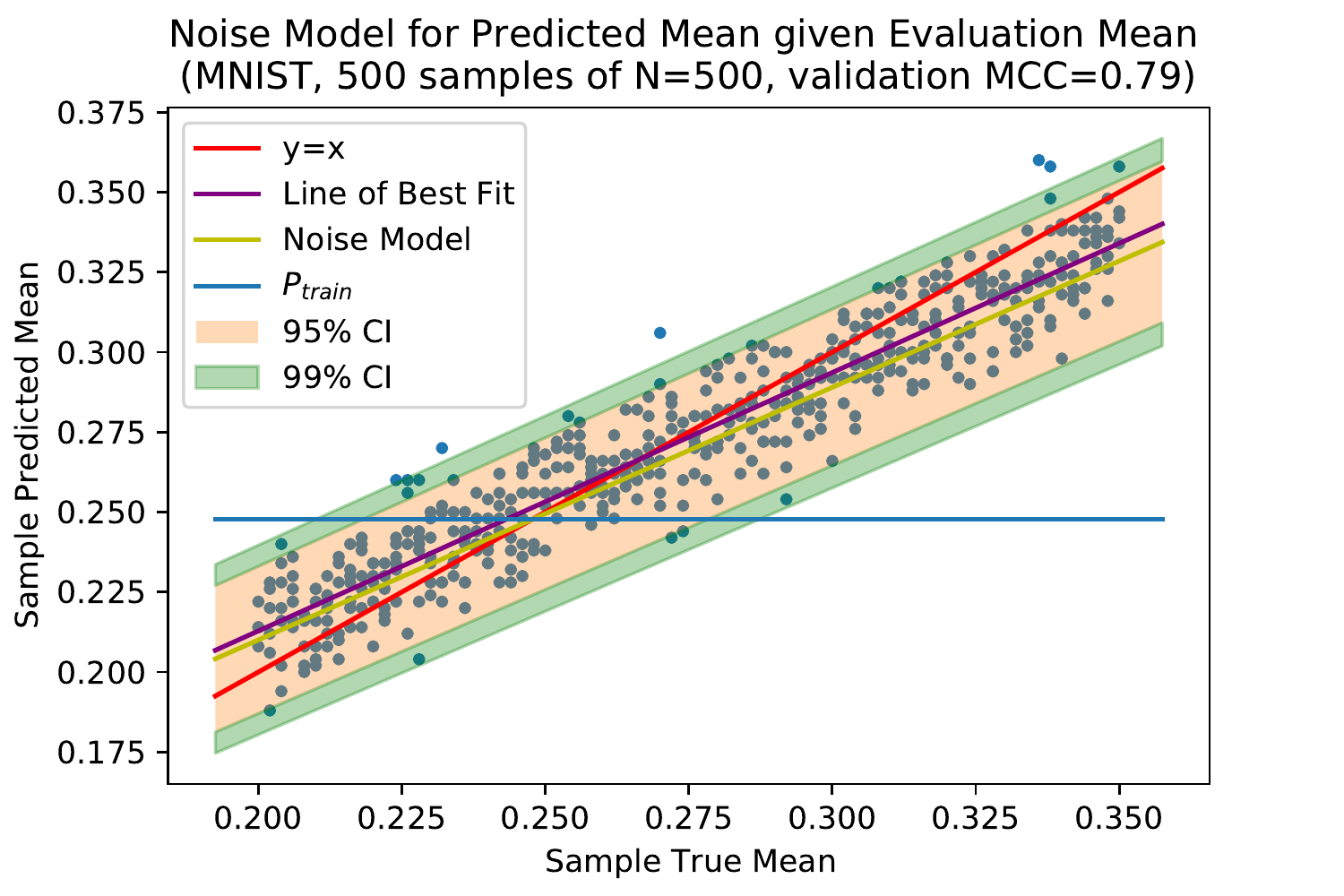}
\caption{Demonstration of noise model success, using an weak model on MNIST, $p_{train}\approx0.25$ and varying $p_{eval}$}
\label{fig:noise_bias_mnist}
\end{figure}

\begin{figure}[htb]
\centering
\includegraphics[width = 0.5\hsize]{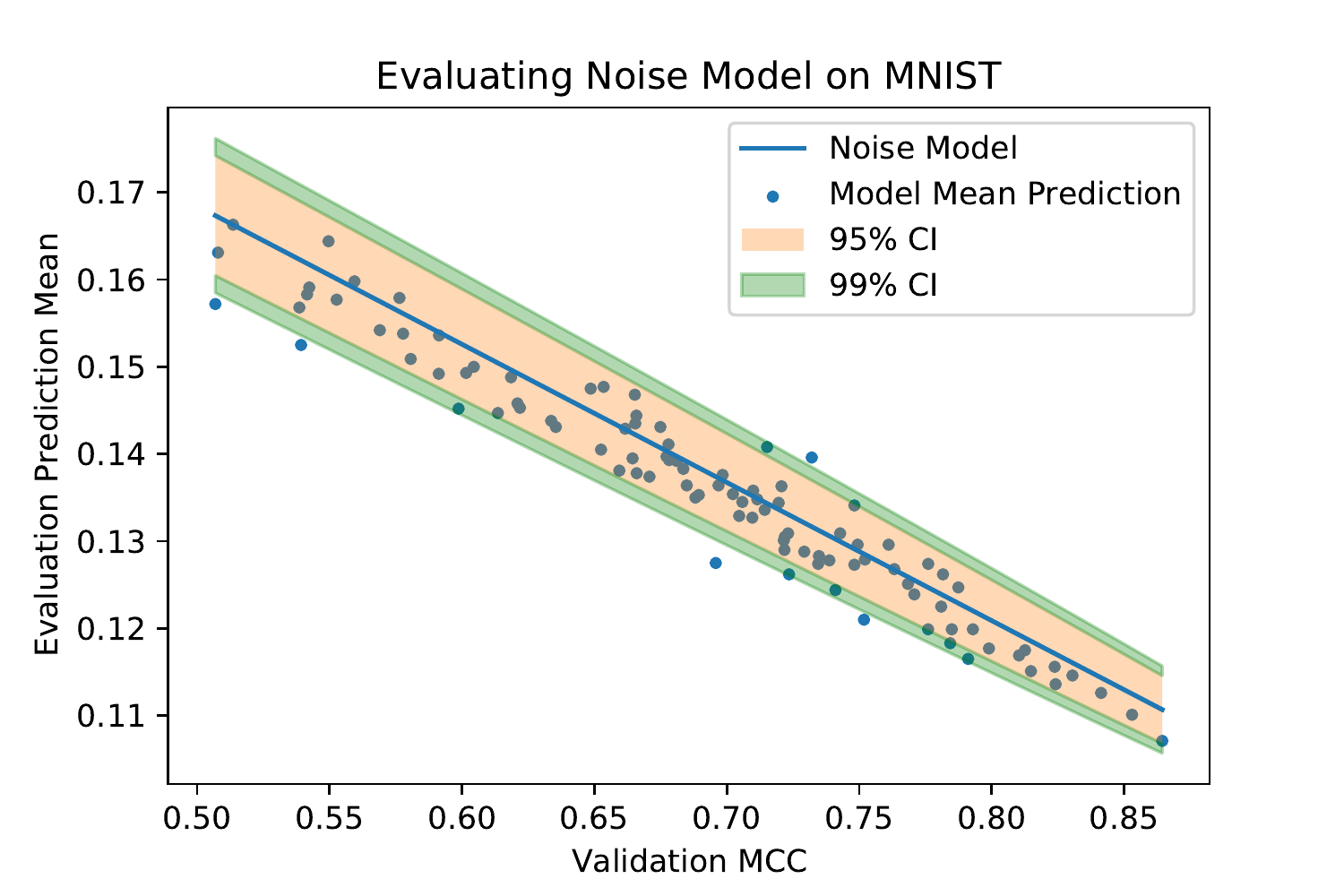}
\caption{Demonstration of noise model success in the same context as \ref{fig:noise_bias_mnist}, keeping constant $p_{train}$ and $p_{eval}$, and varying the Validation MCC. Visualised are model predictions for the evaluation set mean for ten models as they train.}
\label{fig:multiple_models_mnist}
\end{figure}

For evaluation, random subsets of the test set are used as evaluation sets to vary the evaluation set mean. As can be seen in figure \ref{fig:noise_bias_mnist}, the noise model correctly predicts the likelihood of model mean predictions ($\hat\mu$) as a function of the true evaluation set mean ($\mu$). It can also be seen that the noise model expectation (yellow) is extremely close to the Line of Best Fit (purple) and that the expectation of the model's outputs (yellow) is simply an average of the true evaluation set mean (red) and the training set mean (blue), weighted by the MCC (0.79). Additionally, Figure \ref{fig:multiple_models_mnist} shows the success of the noise model in predicting the distribution of predicted means for varying validation MCCs. Here, ten models are trained using the above methodology and paused throughout training to evaluate validation MCC and evaluation set predicted means. One can see the same modelled relationship between MCC, training set mean, and evaluation set as is seen in Figure  \ref{fig:noise_bias} in the theoretical context.\\

This methodology allows for testing hypotheses related to participation bias. Under the null hypothesis, where there is no participation bias, the evaluation set and training set means are identical. Using this model's equation for standard error of the evaluation set mean, one can evaluate the likelihood of mean occurring under the null hypothesis and reject the null hypothesis if this likelihood is sufficiently small. As expected, for a high MCC and    sample size, the model predicted mean on the evaluation set needs to vary less from the training set mean to produce a small p-value compared to a model with a lower MCC. \\

\subsection*{Reversing the Bias}

Having developed a mathematical model for the noise, justified it empirically, and developed a framework for testing for the presence of participation bias, the next step is to reverse it. Doing this is equivalent to obtaining the posterior distribution $P(\mu | \hat \mu)$. Modelling the posterior distribution allows for developing a confidence interval on the true sample mean ($\mu$), rather than merely assessing the likelihood that the evaluation set and training set means are identical. \\

Bayes Theorem implies that: 
\begin{equation}
\PP(\mu|\hat \mu) = 
\frac{\PP(\hat \mu|\mu)\PP(\mu)}{\PP(\hat \mu)}
= \frac{\PP(\hat \mu|\mu)\PP(\mu)}{\int_0^1 {\PP(\hat \mu|\mu) \PP(\mu) \, \mathrm{d} \mu} }
\end{equation}

The likelihood was found above (Equation \ref{eq:likelihood}), for $h$ predicted positives and $t$ predicted negatives:

\begin{equation*}
    \PP(h,t|\mu) = {h+t \choose h} \left(m\mu + (1-m)p_t)\right)^h \left(m(1-\mu) + (1-p_t)(1-m)\right)^t
    \label{eq:likelihood}
\end{equation*}

For $m=1$, this is relatively simple to compute for a prior distribution $\PP(\mu) =  \text{Beta}(\alpha,\beta)$

\begin{align}
    \PP(\mu|\Yhat) \quad\propto&\quad p(\Yhat|\mu)p(\mu) \\
    \propto&\quad \mu^h(1-\mu)^{t}\mu^{\alpha-1}\mu^{\beta-1}\\
    \propto&\quad \mu^{h+\alpha-1}(1-\mu)^{t+\beta-1}\\
    \propto&\quad \text{Beta}(\alpha+h,\beta+t)
\end{align}

For $m=0$, the posterior is simply the prior, as expected. For example, for $\alpha=\beta=1$, the prior is simply the uniform distribution. and the posterior for $m=0$ implies that all values of $\mu$ are equally likely.\\

However, for $m\in(0,1)$, this problem is non-trivial. Given $m$ and $p_{train}$, the likelihood can be re-written as:
\begin{align}
    \PP(\hat\mu|\mu) =& {h+t \choose h} \left(m\mu + (1-m)p_t)\right)^h \left(m(1-\mu) + (1-p_t)(1-m)\right)^t \\
    =& {h+t \choose h}  m^{ht}\left(\mu + m^{-h}(1-m)p_t)\right)^h \left((1-\mu) + m^{-h}(1-p_t)(1-m)\right)^t\\
    =& a(\mu+b)^h(1-\mu+c)^t \label{likelihood_rewrite}
\end{align}

This form does not have a conjugate prior, and hence the marginal likelihood is necessary for numerical computations. For relatively small $n$, the marginal likelihood, and hence the posterior, can be approximately computed numerically. \\

\begin{figure}[htb]
\centering
\includegraphics[width = 0.5\hsize]{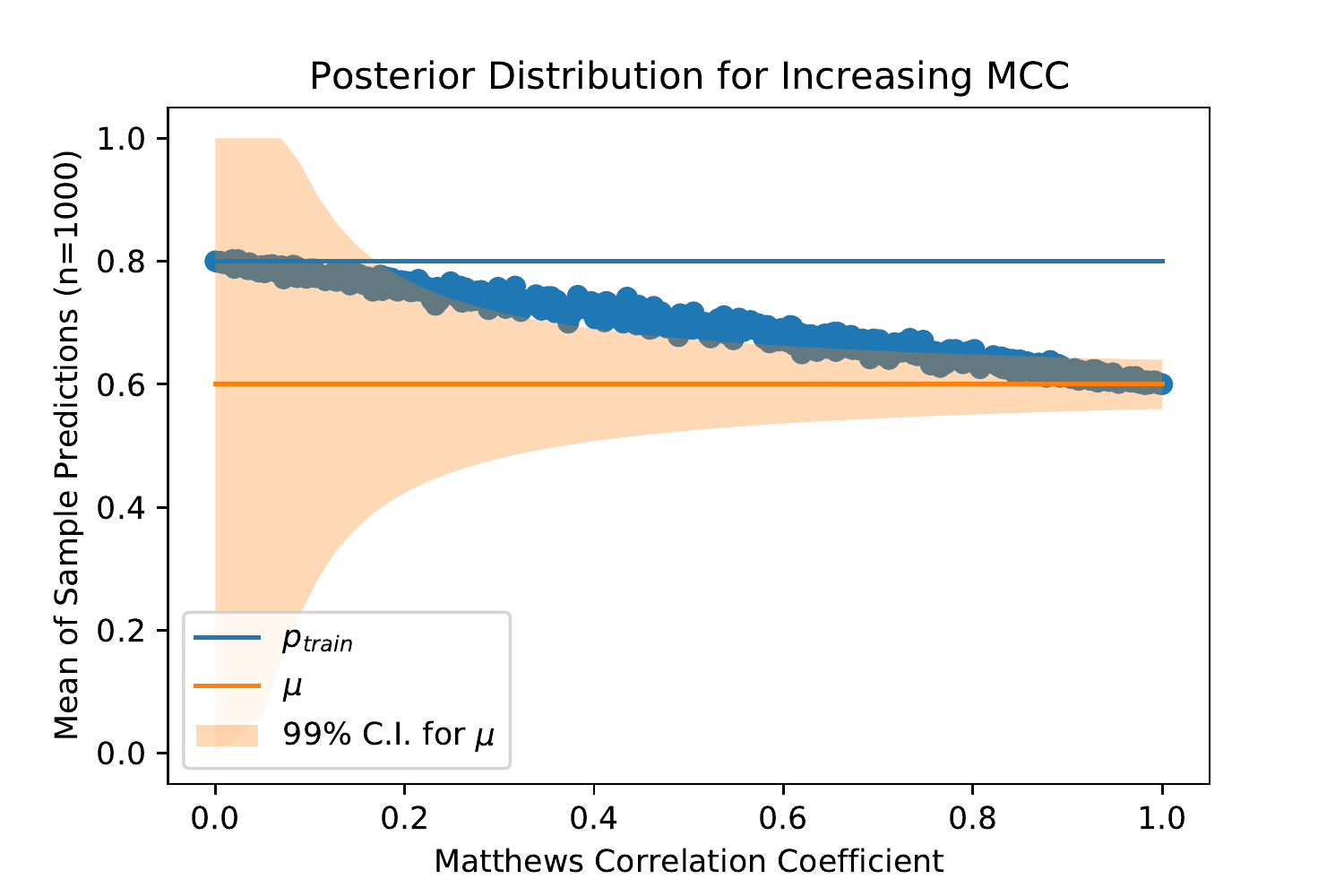}
\caption{Using the posterior distribution of the noise model to predict the true mean of an evaluation set ($\mu$) given the predicted mean ($\hat\mu$}
\label{fig:theoretical_posterior}
\end{figure}

\begin{figure}[htb]
\centering
\includegraphics[width = 0.5\hsize]{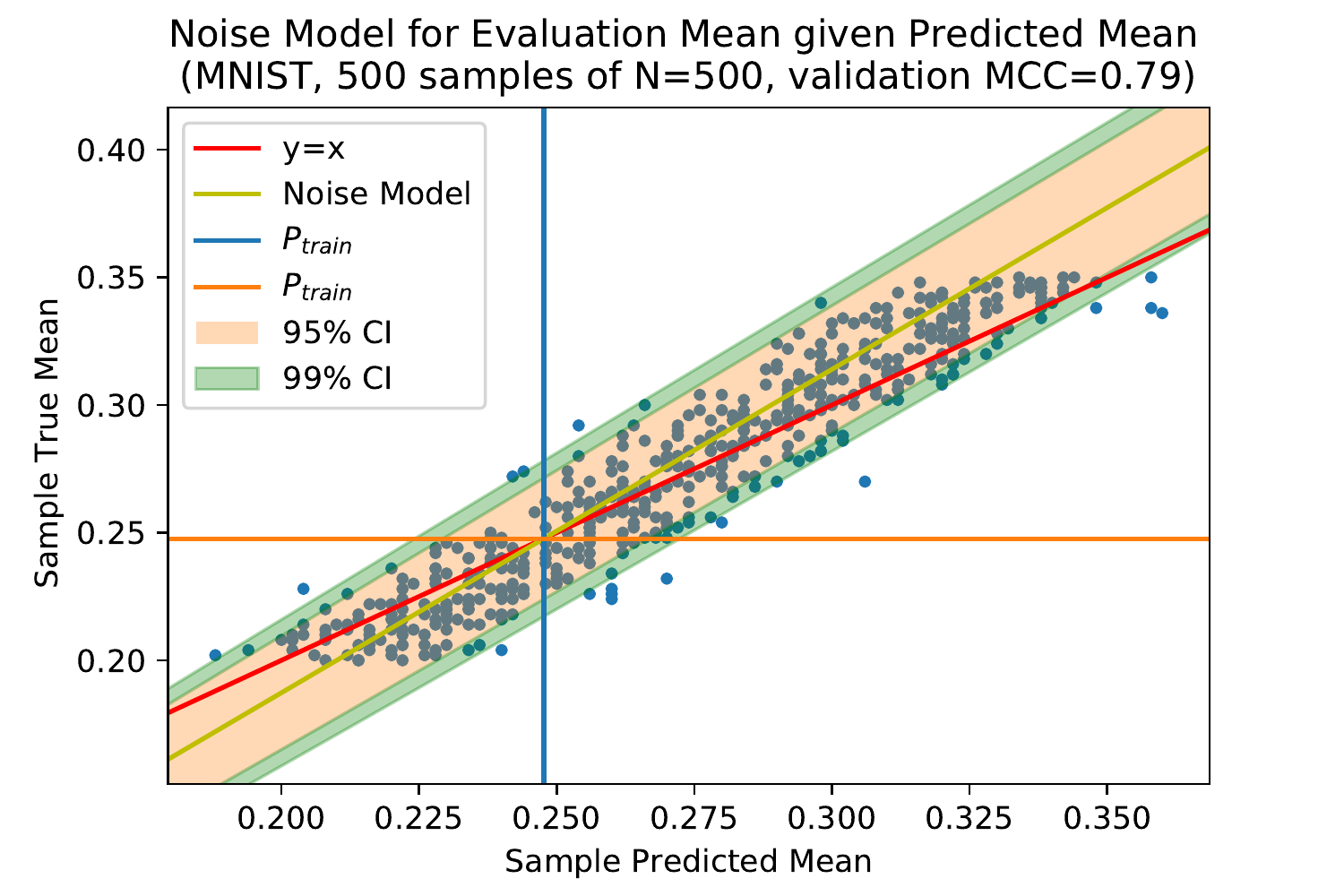}
\caption{Predicting the true mean of an evaluation sample, give the predicted mean, $p_{train}\approx0.25$ and varying $p_{eval}$}
\label{fig:predicting_true_mean_mnist}
\end{figure}

Just as above, one can empirically test this result theoretically as shown in Figure \ref{fig:theoretical_posterior} and empirically on MNIST as shown in Figure \ref{fig:predicting_true_mean_mnist}. The mpmath python library \citep{mpmath} is used to compute integrals numerically due to its ability to work at very high precision.

\subsection*{Discussion}

In this section, a mathematical model is developed for estimating the mean of a binary variable on a given evaluation set given an imperfect Machine Learning model trained on a differently distributed training set. There are, however, limitations to this model that must be discussed. First, Figure \ref{fig:noise_mcc} shows a linear relationship between the degree of noise and the MCC, but not an exact one. For sufficiently large samples, the difference between these two quantities (MCC and 1 - noise) should converge, but for smaller samples, they likely will not. An error term should be added to account for the error in estimating the noise given the validation MCC, which will likely be a function of, at very least, the validation set size. Relatedly, if the training set is extremely unrepresentative of the population from which the evaluation set is taken, the MCC for the validation set and evaluation set very well might significantly differ. For example, for the MNIST example, if the training set had no fives in it at all, or very few, the MCC for the validation and evaluation sets may differ.  \\

Finally, it is worth mentioning that in future work, there are significant improvements that can be made to this model. Other directions were pursued unsuccessfully prior to selecting this model, including the use of probability calibration methods discussion in Chapter \ref{chapter:background}. Under the probability calibration model, model confidences can be treated directly as probabilities, provided that the Expected Calibration Error (ECE) is very low; i.e. $\PP(y_i|\yhat_i)=\B(\yhat_i)$, where $\yhat_i$ refers to single model predicted probability. However, when the ECE is not sufficiently small, it must produce an error term of it's own, i.e. $\PP(y_i|\yhat_i)=\B(\mathcal{D}(\yhat_i))$ for some distribution $\mathcal{D}$. This distribution may be a function of empirical results regarding the ECE, or could even make use of uncertainty quantification using multiple model forward passes with dropout kept on, as described in Chapter \ref{chapter:background}. This methodology may significantly improve this chapter's noise model, but is far more complex than the model developed here, likely requires significant work related to probability calibration, and is therefore left as future work. That said, the model developed thus far has proven quite successful for the purposes of this project, and will generally allow for estimations of participation bias provided that training, validation, and evaluation sets are all large.

\section{Results on Shout Dataset}

In what follows, I will provide a number of results from applying this methodology with the final ToBERT model developed in this project. It is worth noting that confidence intervals are estimates based on the above model and therefore do not account for errors in the model itself. It is possible that when there is \textit{very} significant participation bias, that the confidence intervals may mis-represent the true proportions. Additionally, note that these estimates are based on \textit{number of conversation} rather than number of texters. As described in Chapter \ref{chapter:dataset}, demographic questions about individuals that generally do not vary significantly over the short timescale of the Shout dataset (i.e. age, race, disabilities) and are generally only asked once to texters who respond to the survey multiple times. Therefore, for example, if an individual has three conversations, answers the texter survey once, and in it identifies as female, all three conversations are considered conversations with an individual who identified as female in the texter survey. For questions regarding the conversation itself or how the individual has been feeling very recently, texter survey responses are not extrapolated to other conversations with the same texter. Finally, note that confidence intervals extracted from the posterior distribution are based on a uniform prior. \\

\begin{figure}[htb]
\centering
\includegraphics[width = 0.5\hsize]{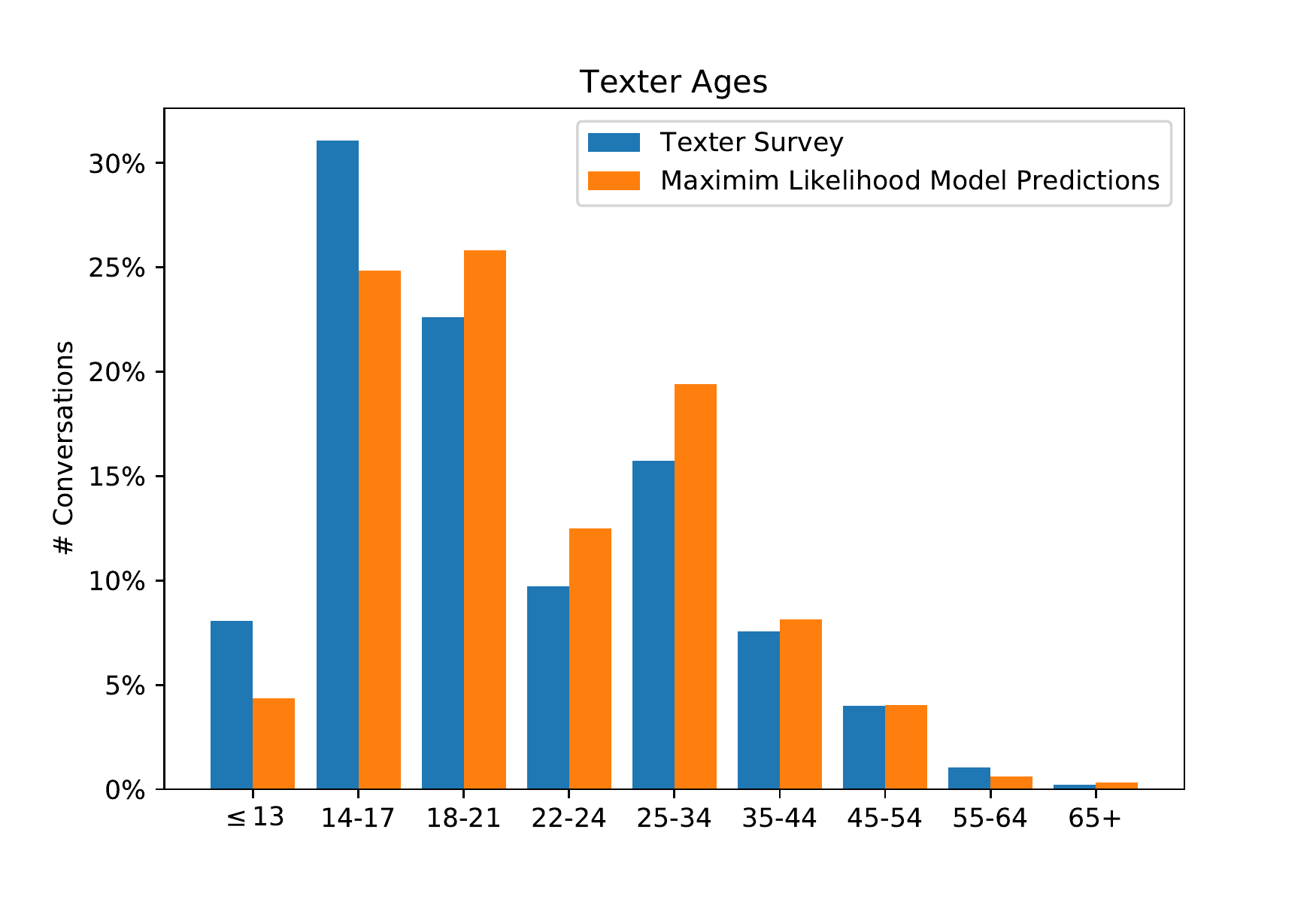}
\caption{Distribution of ages according to the texter survey and according to model predictions. Model predictions shown are maximum likelihood estimates of using ToBERT predictions and the mathematical noise model.}
\label{fig:age_preds}
\end{figure}

To begin, it is likely that age has an impact on survey participation. For example, based on texter survey responses regarding age, 33.87\% of Shout conversations are with individuals under the age of 18. Based on ToBERT model predictions along with the above noise model, for non-respondents, this fraction is in the range , (27.24\%, 27.99\%) based on a 99\% confidence interval (C.I.). This indicates that age has a statistically significant effect on participation, with individuals under 18 significantly more likely to respond to the survey. The overall fraction of Shout texters under 18 is expected to be in the range (99\% C.I.: 28.87\%, 29.53\%). The maximum likelihood model predictions is shown in Figure \ref{fig:age_preds} along with the conversation age distribution from the texter survey. \\

Similarly, females make up 80.31\% of conversations based on the texter survey, but likely only 74.2\% of the Shout conversations are with females (99\% CI: 73.58\%, 74.82\%). Individuals with dyslexia, mobility difficulties, or deafness are all likely underrepresented by the survey\footnote[1]{$p<<0.0001$}. Individuals with ADD/ADHD are likely over-represented, with conversation surveys showing such individuals to make up 7.135\% but model predictions indicating a significantly smaller fraction (99\% C.I.: 2.36\%, 4.49\%). \\

Individuals identifying as ``Heterosexual or Straight" appear under-represented, while individuals identifying as ``Bisexual or Pansexual'' or ``Gay or Lesbian'' are significantly over-represented by the survey*. Individuals identifying by a race with the word ``White'' in it are over-represented by the texter survey, making up 89.419\% of conversations by individuals who fill in the texter survey, compared to an overall fraction of (99\% C.I.: 80.73\%, 83.12\%). Conversely, ``Asian'' races are underrepresented by the survey - 3.881 compared to (4.59\%, 6.36\%). Individuals who served in the military are also likely under-represented*.\\

Finally, the question ``Did you find this conversation helpful?" is subject to very significant participation bias. 85.01\% of survey answers to the question are ``Yes" but model predictions indicate that if every conversation had a corresponding answer, the mean would be in the range (45.45\%, 46.51\%).
\chapter{Measuring Skill}
\label{chapter:measuring_skill}

As discussed in \ref{chapter:dataset}, Shout conversations are typically between texters and trained Crisis Volunteers (CVs). CVs are not always mental health experts, so oversight and review by supervisors and coaches is necessary for ensuring conversations are at a consistent high quality. Conversations are frequently monitored as they take place and a number are reviewed afterwards to provide detailed feedback to CVs to assist with their professional development. Due to practical limitations, only a small subset of conversations are reviewed and at present, the conversations chosen for review are generally randomly selected. Some supervisors have suggested that reviews may be significantly more useful if relevant rather than random conversations were selected for review. \\

This chapter addresses methods of analysing conversations with the machine learning model previously developed to assist in selecting the most relevant conversations for review. Relevant conversations may include those that went particularly well or poorly; conversations where particularly high or low skill was demonstrated; or conversations where a particular type of mistake was made. This chapter will discuss three metrics for conversation success, namely (1) How helpful might the texter have found this conversation; (2) how helpful would one expect that texters generally find this CV, based only on this conversation; (3) how experienced does this CV seem to be, based only on this conversation. These metrics are summarized in Table \ref{table:7results}. In the next chapter, one specific and extremely important mistake CVs can make will be addressed in depth, namely failing to identify properly a texter's suicidal risk. \\

\begin{table}[hbt]
\centering
\begin{tabular}{llll}
 & \begin{tabular}[c]{@{}l@{}}Pearson\\ Corr. Coeff\end{tabular} & \begin{tabular}[c]{@{}l@{}}Explained\\ Variance\end{tabular} & \begin{tabular}[c]{@{}l@{}}\# Ground \\ Truths\end{tabular} \\
\textbf{Texter Reviews:} & 0.562 & 0.219 & 30,085 \\
\textbf{Avg. Texter Review:} & 0.509 & 0.252 & 134,509 \\
\textbf{CV experience} & 0.724 & 0.524 & 233,882
\end{tabular}
\caption{A performance comparison for metrics developed in this chapter, using the ToBERT model. Performance is measured on a hidden test set. For texter reviews, raw model outputs are used without applying the decision function.}
\label{table:7results}
\end{table}

Quantifying how successful a conversation was is not a simple task and is the topic of significant research. Through conversations with Shout coaches, I learned that there are a number key factors that differentiate more and less successful conversations that coaches notice and/or look for. Mistakes CVs make that lead to worse conversations include not asking for or using the texter's name; jumping into problem solving too quickly without exploring the texter's problem sufficiently; or asking too many short questions in an interrogative manner. Factors that make a conversation more successful may include verbally sympathising or empathising with the texter, ending the conversation with a plan for the texter or setting intentions, and the texter successfully de-escalating, i.e. moving from a ``hot'' place to a ``cool'' place. These are just a few examples of factors that CVs look for in differentiated between successful, skilled conversations and less successful conversations. \\

The existence of these key factors indicates that it is possible for an expert to evaluate a conversation, or a CV on the basis of a conversation, based only on the conversation text. Labeling conversations for such errors, however, would be very difficult as it would require significant expert labeling. Additionally, these indicators of successful or unsuccessful conversations cannot be treated as standalone check boxes; checking that each conversation has at least one expression of empathy, at least one exploring question, etc. is neither necessary nor sufficient to ensure conversations are of high quality. Therefore, rather than explicitly searching for these factors, we can therefore focus on simpler robust success metrics that may correlate with these factors and allow coaches, in manually reviewing conversations, to identify the flaws in a conversation and suggest potential improvements. \\

\section{Texter Reviews}

In the texter survey, texters are asked to assess whether they found their conversation helpful, and if so, how helpful on a five point scale. These reviews are useful for assessing a conversation specifically in terms of what makes a conversation feel helpful to texters. These reviews, however, are only available for a small fraction of conversations. As discussed in the previous chapter, the answers to whether the conversations were found helpful were successfully predicted with a reasonably high performance: 0.531 MCC, 0.943 F1 Score and 90\% Accuracy. Whether the corresponding rating was greater than 1, 2, 3, or 4 were predicted with MCC 0.51, 0.49, 0.45, and 0.39 respectively. These predictions can therefore be used as a one conversation success metric. \\

Reviews, however, are likely quite noisy for a number of reasons. The degree to which a conversation is helpful likely has much to do with the texter's state of mind at the time of the conversation and the issue that they were texting about, not merely the CV's skill. For example, it may be the case that texters facing depression give higher or lower reviews, generally, regardless of the nature of the conversation; or it may be that individuals that text about issues that don't benefit from talking tend to find conversations less helpful.\\ 

Additionally, not all texters may understand the questions in the same way or understand the 1-5 scale in the same way, which may explain the significant disparity in MCC between different ratings. This and many of the irrelevant factors that affect a texter's rating are not predictable based on conversation text and are therefore unlikely to bias model predictions as they bias texter ratings. Model predictions, therefore, likely ``average out'' texter biases, predicting how an average texter having this conversation would rate it, rather than the particular texter. The unpredictable noise that affects ratings introduces considerable irreducible error to predictions of texter reviews; as a result, an MCC lower than 1.0 is to be expected - and even wished for - as it indicates that this noise may be being averaged out. \\

\section{Average Texter-CV Reviews}

While model predictions of a texter rating may reduce \textit{some} noise, model predictions will nonetheless remain biased by \textit{predictable} factors related to texter state that may impact decisions. For example, if the type of issue the texter faces significantly impacts texter reviews, the model may look for the conversation topic to predict the texter review, in addition to the factors that actually predict successful conversations (e.g. empathy, exploring). To escape that noise, we can train the model to predict the average reviews that a CV receives based on each individual conversation, rather than the single review corresponding with that conversation itself. This model output then corresponds with the answer to the question "based on this conversation, what kinds of ratings would you think this CV generally gets?" A good answer to this question will pay less attention to the texter's issue and state of mind, as it'll affect reviews, and more attention to the factors that differentiate a CV who consistently receives high reviews and one who consistently receives low reviews. \\

To train a model to answer this question, CVs are simply assigned a ``skill'' rating at each point in time based on their reviews so far. The simplest way to do this is to simply use the average of all reviews up until a conversation as the CV rating at the point in time of that conversation. This is likely to yield the best performance compared to, for example, a weighted moving average of reviews, simply due to sparcity of texter feedback. Rather than using the numerical scale, which may add noise due to interpretation of the scale, only the binary ``Did you find this conversation helpful?'' is used. Additionally, to ensure that this output is not excessively biased by a single texter's review, this label is only used once a CV has received at least five reviews. Additionally, because relative skill or success is the goal rather than the underlying fraction of conversations found helpful, these ratings are finally replaced with their corresponding quantiles. \\

It's important to note that by using texter reviews as a proxy for CV skill or conversation success, it is assumed that an average texter review reflects in some way a CV's abilities. One may argue that the job of a CV is not reflected in the texter survey - it is the job of a CV to bring texters from a ``hot moment to a cool calm'', to use Shout's phrasing, which might not reflect in how helpful they are found by texters. The average texter review metric must therefore be interpreted only as what it is - not a measure of skill per se, but a measure of the texter helpfulness reviews the CV who took this conversation is likely to get. If this is not exactly skill, it may be something related or independently valuable. This measure also assumes that conversations (e.g. type of issue and severity) that each CV handles is generally evenly distributed. Bias can be introduced, for example, if many CVs only work at night and reviews are generally better or worse for nighttime conversations. Relatedly, a source of bias may arise from top CVs and supervisors being referred the most severe conversations, which may have less representative and more harsh reviews. Ultimately, due to the size of the dataset and the number of unique CVs, the effect of this bias is unlikely to be significant. \\

For this metric, 25.2\% of the variance in the ground truth label is accounted for in model prediction. While it's hard to identify the extent of the irreducible error, it is logically unlikely that much more than this amount of the label variation could be modelled. Between two adjacent conversations, the ground truth label (average texter helpfulness rating) for a CV may vary significantly, especially if the CV has recently received just one outlier texter survey review. The level of noise is therefore likely very significant. The explained variance score is not sufficient to demonstrate that this metric is \textit{useful} but is definitely sufficient to show that the model is being trained successfully to predict this metric. \\

\section{CV Level of Experience}
Within the Shout platform, CV experience is indicated by a numerical level based on the number of conversations they've had. As CVs gain experience, they progress through the levels and gain privileges, including being able to participate in multiple simultaneous conversations. CV experience, measured in number of conversations, is therefore implicitly thought of as a proxy for skill and may justifiably be used as such. Two coaches indicated that correlation between CV experience and having successful conversations may be non-linear. New CVs may be more likely to follow the prescribed structure, while intermediate CVs are more likely to venture from that, for better or worse, as they develop their own conversational style. It is worth, therefore, being hesitant in asserting that conversations with more experienced CVs are likely to be better. As experience is currently used as a proxy for skill, it may be valuable to develop a label corresponding with ``based on this conversation, how much experience do you expect that the CV has.'' This label may well accompany the other two previously described in assessing a conversation's success and assessing the CV in a conversation. 

For the ranked experience label, the achieved explained variance score between predictions and labels is 52.4\%. This is much higher than that for average texter responses, which is expected because rank moves far more slowly than average reviews may, and thus the underlying label contains far less noise. However, similarly to the other two labels, a very high coefficient of determination is unlikely desirable. The goal is to identify the likely skill level of a CV in a conversation, not the exact number of conversations that they've had, and therefore failure to perfectly identify the ground truth is to be expected. Again, more is likely necessary to demonstrate the usefulness of this metric. \\ 

\section{Evaluation and Discussion}

This chapter discusses the implementation of three conversation success metrics; evaluating these metrics, however, is very difficult. While accuracy can be measured between model predictions and the corresponding ground truths, all three labels discussed are being trained specifically for noise reduction. As such, perfect accuracy would not be testament to the value of these labels but rather the contrary; even very skilled CVs may occasionally have conversations indicative of a low-skill CV, and likewise CVs with very little experience may occasionally speak in the manner of a very experienced counterpart. Nonetheless, a non-existent correlation between ground truth labels and predicted labels is surely a sign of failure and hence the correlation values presented are at least “sanity checks” for the model training success. \\

The next step in evaluating these metrics is ongoing and is based on their usefulness. Each of metric is being used separately to identify the best and worst conversations from select CVs; these conversations will then be passed on to coaches to read these conversations and provide feedback to the CVs. The coaches will then be able to provide feedback on how useful the metric they used was for identifying ``good'' and ``bad'' conversations. This feedback can be used to improve these metrics and ensure that they are of high quality, for later use with real-time models. \\

\section{Future Work}

One significant limitation in using texter reviews is the limited number of texter reviewed conversations. As mentioned, averages are only taken once a CV has received five reviews, meaning that many conversations are being labeled based on only five or slightly more texter reviews. It is expected that the greater the number of conversations averaged over, the less noise due to any single outlying conversation or review; one solution to this problem is to use model predicted reviews in addition to - or instead of - actual texter reviews. A first model would predict individual texter reviews and a second independent model would then use these model outputs in place of texter reviews. A combination approach may also be used, with the model predicted reviews weighted less. This methodology may produce significantly better results, but requires sufficient success in predicting texter reviews to avoid compounding error. It is therefore deferred to future research, and will be discussed further in Chapter \ref{chapter:discussion}. \\

Additionally, it is worth mentioning that the metrics provided in this chapter are produced from a multi-ask only model. Methods that may lead to improvement of this metric include those described earlier - the use of a larger model and the use of label fine-tuning. As explained in Chapter \ref{chapter:evaluation}, with the CNN-RNN, label fine-tuning for the texter helpfulness rating brought an absolute MCC increase of 0.0385, compared to the multi-task only model. However, the irreducible error for this label may be significant, and thus a much higher metric is unlikely to be achievable or desirable for even a much bigger model. \\  

Finally, a methodology not assessed here that may prove useful involves training a model to predict texter reviews, average reviews, or CV experience \textit{independently} of confounding variables. Confounding variables include conversation topic, time of day, texter demographics, risk of suicide/self-harm, or any other conversational factors that are irrelevant to CV skill but may influence texter reviews or may correlate with a CV's level of experience. A promising way to do this involves adversarial learning in which a model is trained to predict a target without influence from confounding variables. A method for adversarial training can be found in \citet{ganin2015domainadversarial}, in which confounding variables are predicted by the model but trained with inverted gradients so that the model learns the correct task without learning the features predict the confounding variables. \\ 
\chapter{Suicide Risk Ladder}
\label{chapter:suicide_risk_ladder}

Conversations with Shout often involve individuals at significant risk of suicide or self-harm and one of the biggest worries of some coaches and supervisors is conversations where CVs miss risk. In conversations where suicide is mentioned, CVs are instructed to carefully assess suicide risk according to the \textit{suicide risk ladder} composed of four rungs: suicidal desire, intent, capability, and timeframe. The first three rungs are also sometimes called thoughts, plans, and means, respectively. CVs are meant to ask specific and direct questions for each rung, moving up each time they get an answer in the affirmative, while also keeping track of the suicide risk ladder with provided checkboxes on the platform. This process of asking risk assessment questions is known as \textit{laddering up} and is essential for identifying and potentially escalating high-risk conversations to supervisors and for performing an \textit{active rescue} if an individual is at the top of the ladder, where local authorities are asked to intervene in a suicide attempt. A major worry of Shout supervisors is missed risk which can occur when a CV does not ladder up properly in the conversation and thus does not determine the individual's risk or when CV does not report the risk correctly using the provided checkboxes. Identifying conversations where risk was missed in hindsight may be very useful in providing feedback to CVs, helping CVs improve in their ability to risk assess. \\

In the previous chapters, the developed Machine Learning model is trained on the suicide risk ladder during multi-task training, and it therefore is likely capable of identifying when suicide risk is present and where errors are made in ticking the correct boxes. It may, however, have an additional function beyond identifying risk - it may be able to assess a CV's success in laddering up within the conversation correctly. \\

In CV's educational literature, CVs are trained to ask laddering up questions clearly and to clarify where texters are unclear. In one example conversation, the CV asks if the texter has thought about suicide and the texter responds "I dunno. Kind of." The example CV responds "It takes courage to talk about this. Just to clarify, do you mean you want to end your life?" In conversations with clear and successful laddering up, there will generally be little uncertainty as to the individual's place on the risk ladder; whereas where the CV moves on from an indication of risk without asking clarifying questions, uncertainty will be high. Model uncertainty therefore might be a useful metric for assessing how well a CV performed laddering up. \\

\section{Metrics}

\begin{table}[]
\begin{tabular}{lllll}
                            & \textbf{F1 Score} & \textbf{AUC-ROC} & \textbf{Avg Precision} & \textbf{MCC} \\
\textbf{Desire (thoughts)}    & 0.834             & 0.955            & 0.895                  & 0.752        \\
\textbf{Intent (plan)}      & 0.837             & 0.978            & 0.898                  & 0.797        \\
\textbf{Capability (means)} & 0.814             & 0.981            & 0.871                  & 0.783        \\
\textbf{Timeframe}          & 0.774             & 0.985            & 0.824                  & 0.756       
\end{tabular}
\caption{Results on final Transformer Over BERT model on Suicide Risk Ladder labels}
\label{table:results_suicide_risk_ladder}
\end{table}

Results from evaluating the ToBERT model on Suicide Risk Ladder questions can be seen in \ref{table:results_suicide_risk_ladder}. Upon initial review, these metrics might either seem very high, relative to the difficulty of the task at hand, or unacceptably low, given the cost of errors in assessing risk. Due to the quantity of data and complexity of the ToBERT model, it is unlikely that significant improvements upon these metrics can come from Machine Learning solutions. Admittedly, the model used is considered a ``small'' model, on the scale or state-of-the-art deep learning, and for these metrics, label fine-tuning did not take place. Using a larger model and fine-tuning on the relevant labels are likely to very slightly improve metrics, but not too significantly. Additionally, the quantity of data used was quite large, so it's unlikely that the errors are due to a lack of data. These facts indicate that a proper error analysis is warranted. The likely explanation for the relatively low performance is irreducible error.\\

\section{Error Analysis}

The likely high irreducible error warrants a proper error analysis. To this end, Shout was able to provide a re-assessment of suicide risk by experts, rather than crisis volunteers such as those who answered the survey. Focusing primarily on the suicidal timeframe question, twenty conversations were selected including ten randomly sampled conversations where the CV and AI agreed on the presence of a suicidal timeframe and ten conversations with disagreement. Two reviewers were asked to indepedantly annotate each conversation according to suicidal risk exactly as a CV would at the end of the conversation. This review methodology was imperfectly completed by the reviewers, with conversations only reviewed by one individual. Hence reviewer subjectivity or errors were not appropriately accounted for. Additionally, due to the design of the Shout platform, it was impractical to avoid the reviewers seeing the original labels provided by the CVs, introducing a source of bias. \\

As stated, of the 20 conversations, the AI model agreed with 10 of the original CV labels for the presence of suicidal timeframe and disagreed with 10. Using the expert labels from the reviewers, the agreement rate increased drastically to 15 of the 20 conversations. Equally interestingly, the original CV and the reviewer only agreed on 9/20 labels. A sub-sample of these conversations were also re-reviewed in a video chat with myself and the two reviewers. In four conversations, capability was labelled as a `no' by the reviewer, but contained explicit plans that seemed to strongly indicate a capability to complete the plans (e.g. `rope' or `paracetamol'). Another conversation was marked positive for thoughts and means and not plans though the texter wrote that they have a specific means and they want to use it - which one might think naturally implies a plan. Due to privacy concerns, redacted extracts from these conversations were not able to be provided. In bringing these discrepancies to the attention of the reviewers, some of these discrepancies were explained as errors in the review; while these may have been simple mistakes in ticking the correct box, it is also entirely possible that the reviewers may have missed indicators in the conversation. \\

It also may have been the case that the coaches reviewing these conversations were more concerned with precision than recall, i.e. they did not want to label a conversation as containing a certain level of risk if they were not fully certain that that risk was present. Generally, assessing suicide risk requires very clear and explicit questions, as explained above and in the training literature; the exception, however, is in cases in which the risk at a certain level was made clear by the texter without needing to be questioned, in which case the CV is expected to continue the risk assessment from the next rung of the ladder. For example, if the texter volunteers that they have a plan, the CV should ask about means and timeframe, but need not ask about thoughts. Based on explanations of their reviews, it seemed that coaches looked specifically for correct procedure in assessing risk; as such, where the CV did not ask specific questions following the risk ladder but risk was clearly indicated by the texter, coaches were more likely to miss the indications of risk.\\

\section{Discussion}

Following this review, there are a number of considerations relevant for error analysis. First, model performance is likely considerably higher than indicated by the test set metrics, as shown by the significantly increased model performance on the reviewer's annotations compared to the CVs. One reason for this is that human CVs and even expert reviewers may be more prone to oversight than an AI. Reviewers have limited time and attention capabilities and were generally looking for the laddering up stages in a conversation; if this laddering up did not follow the format that they expected or was not performed correctly, they were more likely to miss risk that was indeed present. By contrast, the AI is trained with neural dropout, where conversation content and internal states are hidden from the model during training, ensuring it learns a robust algorithm that takes into account the entire conversation in decision making. The effect of this is that model outputs are likely to reduce noise in underlying labels, due to the model's robust training and consistent attention given to the full conversation. \\

Second, there is likely significant subjectivity present in determining suicide risk at a given level. CVs must determine where to start laddering up based on the risk initially volunteered by the texter; determining the risk indicated by the texter and whether clarification is needed may involve a degree of subjectivity, with different experts interpreting a texter's message differently. This subjectivity may perhaps decrease with experience level, with more experienced CVs agreeing more often with each other than less experienced CVs would. Beyond subjectivity, an intrinsic disparity may exist between CVs and supervisors in how they assess risk. The coaches I spoke with were more worried about false positives than false negatives, but CVs I spoke with expressed the opposite concern. \\

Model outputs may therefore ``average out'' subjectivity, outputting what an average CV would predict on a given conversation. While that may be somewhat desirable, as it would lead to greater consistency, the average CV label may not correspond well with a ``gold standard'' risk assessment. In future work, an actual set of gold standard risk assessments may be fine-tuned to produce model outputs less biased by the average CV and more inclined towards more desirable outputs. Alternatively, risk assessments may be trained with a weighted loss function, taking into account more experienced CV assessments more than newer CVs, under the assumption that experienced CVs label more closely to a ``gold standard''. \\

\section{Uncertainty Quantification}
One clear result of the error analysis is lack of clarity in suicide risk is an indication of unsuccessful laddering up. As explained earlier and in training literature, risk assessment questions are expected to be specific and involve follow up to ensure clarity to the best extent possible; performing laddering up risk assessment well in a conversation is likely to bring about high certainty in the texter's suicidal risk at a given rung compared to less successful laddering up. A practical application of this is that model uncertainty can be used as a label for whether sufficient risk assessment was present in the conversations. \\

To assess this claim, a second stratified set of conversations were sent for review by five reviewers. A larger sample size of 100 conversations was sent, including 25 true positives, true negatives, false positives, and false negatives, based on the original CV label. Reviewers were asked to reassess suicide risk, as in the earlier review, and also to assess whether the CV ``laddered up'' was satisfactorily. For example, laddering up is to be considering satisfactory if the topic did not discuss suicide and the CV did not ask any laddering up questions; or if, say, suicidal thoughts were volunteered and the CV proceeded to ask specific laddering up questions beginning with those about a plan. Inversely, laddering up is unsatisfactory if suicidal thoughts are volunteered but no questions are asked about plans; if questions are asked but aren't sufficiently clear; if unclear answers aren't followed with clarifying questions; or if a rung of the ladder is skipped, despite the relevant information not being volunteered by the texter. \\

Reviewer results were not received in time for inclusion in this report, but will answer key questions in future. First, this review is sufficiently large to generate statistically significant results, and can thus be used for obtaining more robust metrics for model performance in suicide risk assessment. Based on results from the first review, these results are likely to indicate that the model's success is in spite of noise, and that its performance relative to a gold standard is significantly better. \\

Additionally, the two methods for model uncertainty quantification in Chapter \ref{chapter:background} can each be assessed for their ability to predict unsatisfactory risk assessments. The first method used is dropout uncertainty quantification, in which ten model outputs are produced for each example with 10\% dropout kept on at evaluation time; the standard deviation of these outputs is used as a quantification of uncertainty. The second method is to calibrate probabilities using the Logistical Regression method described in Chapter \ref{chapter:background} and then use the model prediction distance (without dropout) from 0.5 as a measure of model certainty. Each of these methods will be evaluated by Average Precision, to identify how well they separate between conversations found to have satisfactory and unsatisfactory risk assessments. \\ 
\chapter{Discussion}
\label{chapter:discussion}

In this project, increasingly complex models are produced to read conversations between anonymous texters and the Shout crisis text-line and generate a number of predictions. The final model, an end-to-end Multi-Task Transformer over BERT model, is shown to produce strong performance, even with a relatively small model in a multi-task-only setting. A mathematical model is presented for reversing participation bias from a texter survey using the model; three metrics for conversation success are developed and assessed; and the applicability of the model for identifying suicidal risk and assessing risk assessments is shown. However. the work presented in this paper is an exploration, and as such, there is no shortage of further work to complement that which is presented here. \\

This chapter will address some of the research to follow the conclusion of this project. These include potential improvements to the machine learning model developed here for supervised training on the Shout dataset. There are also more general improvements to the methodologies developed, including training with hierarchical transformers, multi-task training with auxiliary labels, and the mathematical model for reversing biased noise. There further mental health related research to conduct on the basis of model outputs. And, importantly, there is significant room for real-time application of this methodology using using Knowledge Transfer. \\

\section{Future Work}

There are a number of potential improvements to the model developed in this project that may lead to increased performance in future work. As discussed in Chapter \ref{chapter:evaluation}, the use of multi-task-only training produces slightly decreased performance compared to label fine-tuning following multi-task pre-training. The solution to this may involve the use of a much bigger BERT model, fine-tuning on individual labels, or fine-tuning on subsets of the most relevant labels. Relatedly, many papers discussing BERT fine-tuning, e.g. \citet{roberta}, indicate the importance of hyper-parameter tuning, which is only very slightly performed here; with greater time and resources, tuning of learning rate, weight decay, dropout rate, and batch size is likely to improve model performance. Additionally, extremely recent alternatives to ToBERT may be prove more efficient in obtaining similar performance metrics, such as the Longformer \citep{longformer} and Big Bird \citep{bigbird}. \\

A significant change that could improve performance involves adding more data to the model inputs. The Basic Features model is trained on length and time-related features and shows a small amount of success on a number of labels, including texter helpfulness reviews. The time-related features used here are not made available to any of the later models, including the final ToBERT model, despite their apparent usefulness in predicting certain labels. In future work, modifications to the model may be made to allow for message timestamps to be inputted alongside messages, which would provide signal not available in conversation text. This could provide valuable information such as the rate at which the texter and CV are typing and responding, significant pauses that occur between messages, and the duration of wait time at the start of the conversations. \\

Chapter \ref{chapter:participation_bias} presents a novel mathematical model for identifying and reversing training set bias. Future research is likely to yield various interesting results regarding Shout's demographic distribution and any number of conditional distributions (e.g. How many individuals in a certain age range of a certain gender would say that they ``can't stop worrying over the last 2 weeks''?). 
There are also a number of improvements to the mathematical model developed in the chapter, such as accounting for model confidence and uncertainty in more precisely obtaining the posterior distribution. Additionally, further empirical evidence may be helpful to better understand the reliability of noise model in alternative situations; for example, to better understand the types of machine learning models for which the mathematical model is applicable. \\

Chapter \ref{chapter:measuring_skill} develops three metrics from assessing conversation success through the context of CV skill. As discussed there, further work is currently underway to demonstrate practical usefulness of each of these three metrics. A number of alternations may lead to even more robust metrics in the future. These including the use of separate model predicted texter reviews in the absence of real texter reviews, and the use of adversarial training to explicitly reduce noise from confounding variables. \\

The metrics developed may also be used for better understanding the factors that contribute to successful Shout conversations and differentiate the best of CVs. The degree to which the model pays attention to specific tokens can be analysed to understand the impact of specific words on having a conversation predicted to be successful. Similarly, the relative importance of CV messages and texter messages to a successful conversation may be analysed. To do this, the impact of a given token or set of tokens on a model prediction can be simply quantified by obtained the gradient of the model's prediction with respect to each token. Relatedly, by labelling conversations by the five stages mentioned in Chapter \ref{chapter:dataset}, the relative importance of each stage could be assessed. And if a gold standard annotation of conversation success were produced in the future, for example in terms of whether a texter's state of mind was successfully de-escalated, this labelling could be compared with the three metrics developed here, and could potentially even be predicted successfully by fine-tuning these metrics on the gold standard labels. \\

Chapter \ref{chapter:suicide_risk_ladder} discusses the model's ability to predict suicide risk on Shout's suicide risk ladder, as well as the use of model uncertainty to identify unsatisfactory risk assessments. As discussed, an expert labelling of suicide risk and laddering up in a subset conversations is ongoing, and is expected to demonstrate higher performance on assessing suicide risk than is shown based on metrics from the noisy given labels. This expert review will also indicate the degree to which model uncertainty reflects a failure in laddering up. Improvements, as discussed, will likely include the use of a weighted loss function to emphasise ``gold standard'' labels over noisy labels. Additionally, fine-tuning a decision boundary may be beneficial in improving precision or recall at the cost of the other, based the degree to which a false positive or false negative is considered problematic.

\section{Real-Time Modelling (GPT-2)}

The models developed in this paper aimed to perform as well as possible on conversations that had ended. As such, the models trained in this project were all bi-directional allowing for use of right-to-left dependencies, in addition to chronological dependencies. As explained in Chapter \ref{chapter:background}, the use of bi-directional features can significantly improve model performance, as empirically demonstrated in \citet{bidirectionalrnns}. However, many practical applications of the technology developed in this project require the use of real-time predictions, where features from the end of a conversation obviously cannot be used. A real-time model may be useful for a large number of tasks related to this project; to give a few examples of applications: Predicting suicide risk as a conversation takes place can be used to remind the CV to perform laddering up, or high likelihood of risk in all four rungs can push an alert for supervisors to get involved. Similarly, conversation success prediction can be used to identify conversations that are not going well and to bring them to the attention of supervisors. And real-time predictions of the age of a texter may better inform CVs in providing texters with resources relevant to them. \\

To make these predictions in real-time, a unidirectional model is needed. One model which is very comparable to BERT but uni-directional is GPT-2 \citep{gpt2}, which has a very similar transformer architecture to BERT and the same tokenizer as RoBERTa. The difference with GPT-2 is that it is pre-trained on an auto-regressive language modelling task, rather than masked language modelling task, in which it predicts the next token in a sequence given the preceding tokens. Early in this project, GPT-2 was successfully trained with the language modelling task on the Shout dataset and used as a generative model as a demonstration of the capability of deep neural networks to understand crisis text conversations. Even without fine-tuning on a supervised task, GPT-2 may be useful for allowing new CVs to practice conversations with AI-driven texters; and perhaps even for providing suggested replies to CVs, to improve typing efficiency and to help when a CV is not sure what to write. \\

Following pre-training, GPT-2 model can be fine-tuned much like BERT for supervised learning; unlike BERT, however, GPT-2 can generate label inferences given only the start of a sequence. GPT-2 will not achieve as strong performance as BERT where a full sequence is available due to the lack of bi-directional features; nonetheless, it is known to perform nearly as well and thus can provide significant utility in real-time settings. Additionally, because real-time predictions are generally most useful relatively early in the conversation, the use of hierarchical transformers or a different document-level classification technique would not be necessary - predictions can simply be generated only for the beginning of a conversation, until the maximum sequence length of GPT-2 is reached. 
The practical application of this is that GPT-2 can be fine-tuned to make the same types of predictions as the ToBERT model in this project, but in real-time, and likely with lower accuracy. \\

\subsection*{Knowledge Transfer}

Fine-tuning GPT-2 in the simple way described is unlikely to provide as high performance as is achieved by ToBERT, especially due to the presence of label noise. In the presence of noise, ToBERT, being a more powerful model and being trained on an entire conversation, will more easily develop a robustness to noise. By contrast, with GPT-2, the model is tasked with predicting a label throughout the conversation, even where the most relevant signal from conversation may not occur until late in the conversation. \\

To better explain this problem, consider a conversation where the texter's age is stated explicitly towards the end of a conversation, but without a corresponding texter survey for the conversation, the ground truth label for the texter's age is absent. ToBERT is able to use the entire conversation when making predictions, and as such predicts age with very high performance. Even if not trained on age for that conversation, ToBERT will likely predict other labels based partially on age, and the gradients from those labels will therefore improve its ability to recognise age. In fact, ToBERT achieves an 88.4\% test set accuracy on in predicting whether a texter is over 21 in conversations\footnote{based on ground truths from the texter survey}, though they seldom contain explicit ages. It is therefore very likely that for a conversation where an age is explicitly stated, ToBERT will output correct age predictions with high confidence, even in the absence of a ground truth for the conversation. By contrast, most of the predictions from GPT-2 will be from before the age prediction is stated explicitly, and so the model will have to learn predict the texter's age based on other features, like word choice, style, or conversation topic. GPT-2's training may thus be significantly less robust to noise and label sparcity compared to ToBERT. \\

A solution to this problem is to use the ToBERT model for de-noising and for filling in any empty labels. More specifically, ToBERT can be used for Knowledge Transfer, acting as a ``teacher'' model for GPT-2. Knowledge Transfer is described in Chapter \ref{chapter:background} in reference to knowledge distillation, and involves training the student network to predict the full output distribution from the teacher network, not merely the model's prediction. In this context, it would involve training the GPT-2 model to have its output at the end of each message of a conversation match the output distribution from the ToBERT model applied to the entire conversation. \\

Predicting ToBERT's output distribution enables accounting for model confidence as well as predictions, with more confident ToBERT predictions providing a greater loss, and penalising GPT-2 for having overly confident predictions when ToBERT's confidence is low. Accounting for confidence is essential because for many labels, ToBERT has much higher performance on examples where its confidence is high. Using the ``Age $\leq$ 21'' label, while test accuracy is 88.4\%, looking only at the 50\% of predictions that are most confident (i.e. 25\% of the lowest outputs and 25\% of the highest outputs), test set accuracy increases drastically from 88.4\% to 98.2\%. And similarly, selecting only the most confident 20\% of predictions, ToBERT achieves a perfect 100\% accuracy on the test set. \\

Significant further exploration may be necessary to achieve a production-ready real-time model. To avoid compounding errors, ToBERT outputs may need to be biased by ground truth labels, where they're present, for example by taking the weighted mean between ToBERT's output and the ground truth label. Additionally, significant experimentation and assessment of the model will be necessary to ensure that the predictions are of sufficiently high quality for use in such a sensitive context. \\

% \nocite{*}

%% bibliography
\bibliography{references}

% Should be apa

\end{document}